\documentclass[lettersize,journal]{IEEEtran}
\usepackage{amsmath,amsfonts}
\usepackage{algorithmic}
\usepackage{algorithm}
\usepackage{array}
\usepackage[caption=false,font=normalsize,labelfont=sf,textfont=sf]{subfig}
\usepackage{textcomp}
\usepackage{stfloats}
\usepackage{url}
\usepackage{verbatim}
\usepackage{graphicx}
\usepackage{cite}
\usepackage{hyperref}
\usepackage[numbers]{natbib}
\usepackage{inconsolata}
\usepackage{comment}
\usepackage{graphicx}
\usepackage{svg}
\usepackage{adjustbox}
\usepackage{colortbl}
\usepackage{lmodern}
\usepackage{amssymb}
\usepackage{caption}
\newtheorem{theorem}{Theorem}

\newtheorem{proof}{Proof}[section]
\newtheorem{definition}{Definition}

\begin{document}

\title{ZeroLM: Data-Free Transformer Architecture Search for Language Models}


\author{Zhen-Song Chen,~\IEEEmembership{Senior Member,~IEEE,} Hong-Wei Ding, Xian-Jia Wang, Witold Pedrycz~\IEEEmembership{Life Fellow,~IEEE}
        
\thanks{This work was supported in part by the National Natural Science Foundation of China under Grants 72171182 and 72031009. (\textit{Corresponding authors: Hong-Wei Ding and Xian-Jia Wang})}
\thanks{Z.-S.~Chen is with School of Civil Engineering, Wuhan University, Wuhan 430072, China and Faculty of Business, City University of Macau, 999078, Macao Special Administrative Region of China (e-mail: zschen@whu.edu.cn).}
\thanks{H. W. Ding is with School of Civil Engineering, Wuhan University, Wuhan 430072, China (e-mail: hwding606@whu.edu.cn).}
\thanks{X. J. Wang is with Economic and Management School, Wuhan University, Wuhan 430071, China (e-mail: wangxj@whu.edu.cn).}
\thanks{W. Pedrycz is with Department of Electrical and Computer Engineering, University of Alberta, Edmonton, AB T6G 2R3, Canada and Research Center of Performance and Productivity Analysis, Istinye University, Istanbul, Türkiye (e-mail: wpedrycz@ualberta.ca).}


}


\maketitle

\begin{abstract}
Neural architecture search (NAS) provides a systematic framework for automating the design of neural network architectures, yet its widespread adoption is hindered by prohibitive computational requirements. Existing zero-cost proxy methods, while reducing search overhead, demonstrate inadequate performance in architecture ranking tasks, particularly for Transformer-based models where they often underperform simple parameter counting metrics. Current automated proxy discovery approaches suffer from extended search times, susceptibility to data overfitting, and structural complexity. This paper introduces a novel zero-cost proxy methodology that quantifies model capacity through efficient weight statistics computation while decomposing Transformer architectures into functionally distinct sub-modules, thereby optimizing the balance of their contributions to overall performance. Our comprehensive evaluation demonstrates the superiority of this approach, achieving a Spearman's rho of 0.76 and Kendall's tau of 0.53 on the FlexiBERT benchmark. The proposed method exhibits exceptional computational efficiency while maintaining robust performance across diverse NAS benchmark tasks, offering a practical solution for large-scale architecture search.
\end{abstract}


\section{Introduction}
In recent years, deep learning has proliferated across diverse domains. Traditional neural network design has relied heavily on expert knowledge and trial-and-error processes~\cite{krizhevsky2012imagenet,he2016deep}. As performance demands have increased, the manual design of efficient neural architectures has become increasingly challenging. Neural architecture search (NAS) emerged to address this issue, though early approaches required substantial computational resources, primarily employing reinforcement learning (RL)-based search methods~\cite{zoph2016neural}.
    \begin{figure}[ht!]
            \centering
            \includegraphics[width=0.5\textwidth]{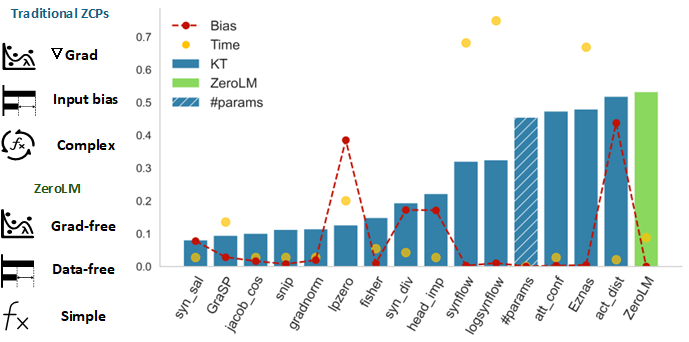}
            \caption{Traditional ZCP vs. ZeroLM in terms of ranking ability, data bias and time cost.}
            \label{A1}
    \end{figure}
The automation of neural network design has proven effective, prompting researchers to shift focus from architecture design to NAS algorithm development. Early NAS methods employed RL for training, discovery, and evaluation. These systems are effective but time-consuming and resource-intensive. \citet{pham2018efficient} introduced weight-sharing supernet enabling one-shot NAS, while \citet{liu2018darts} developed a differentiable approach to transforming discrete problems into continuous ones. These methods, however, still required evaluation of real-world tasks to identify optimal architectures. Zero-shot NAS emerged to reduce training and evaluation costs through zero-cost proxies (ZCPs), which are training-free and obtain proxy scores by randomly initializing architectures and processing mini-batch data. \citet{mellor2021neural} and \citet{abdelfattah2021zero} focused on computer vision applications, while \citet{zhou2022training} adapted ZCPs for Transformer architecture search. Pruning-based metrics have also served as zero-shot NAS proxies. Several limitations persist: hand-crafted methods and those adapted from computer vision often prove unsuitable for Transformer architecture search; automatically designed ZCPs incur significant development costs; and while existing methods achieve good performance, they still require mini-batch data for evaluation. Moreover, automatically designed ZCPs typically combine multiple metrics, introducing additional computational overhead \cite{akhauri2022eznas,dong2024lpzero}. 
The number of parameters for Transformer-based model search have demonstrated promising performance with minimal computational cost. The current challenge remains: ``\textit{How can we improve the performance of zero-shot NAS for Transformer-based models}?''

In this work, we focus primarily on Transformer-based model search. The performance of these models correlates strongly with their parametric scale, with scaling laws \cite{kaplan2020scaling} demonstrating that large language model (LLM) performance relates sublinearly to number of parameters. We hypothesize that novel data-free metrics can be derived as ZCPs from the inherent properties of Transformer architectures themselves. Furthermore, we propose decoupling the Attention and Feed-Forward Network (FFN) modules within Transformers, as these components contribute differentially to overall model performance.

Our proposed method addresses evaluation errors from data input by analyzing solely the model's static architecture, calculating and scoring architectural changes. This new metric eliminates task biases stemming from data influence. By optimizing parameter balance, we efficiently identify ZCPs suitable for the current NAS task. Unlike automatic ZCP searches, our approach reduces sample evaluations during searching, with zero-shot or one-shot hyperparameter optimization enabling task adaptation.


We summarize our main contributions as follows:
\begin{itemize}
    \item We propose a novel ZCP for Transformer-based model NAS that operates without data input, offering rapid computation, high efficiency, and superior performance, as shown in the Figure~\ref{A1}.
    \item Unlike automatic proxy search methods, our approach requires at most one evaluation to optimize parameters and enhance performance for specific search tasks.
    \item Our proxy demonstrates strong correlation across multiple Transformer-based model benchmarks, outperforming baseline proxies. We also analyze the proxy's differential behavior across various model architectures.
\end{itemize}

The rest of this paper is organized as follows: Section \ref{sec2} provides a brief review of existing methods for NAS. Section \ref{sec3} presents a ZCP model. Section \ref{sec4} presents and analyzes the experimental results. Section \ref{sec5} conducts the ablation study. Finally, Section \ref{sec6} concludes this paper.

\section{Related Works}\label{sec2}

\subsection{Zero-shot NAS}

\citet{zoph2016neural} introduced an effective but resource-intensive RL-based NAS. \citet{zoph2018learning} utilized 500 GPUs to search for an efficient image recognition model. To address training inefficiency, \citet{pham2018efficient} developed a one-shot NAS approach. Building on this, \citet{liu2018darts} proposed a differentiable method using gradient descent for optimization. \citet{chen2021neural} introduced a training-free approach employing the NTK method to estimate neural network performance. \citet{lin2021zen} evaluated models using input image perturbation norm, while \citet{mellor2021neural} calculated activation distances through kernel functions after mini-batch processing. Various pruning metrics have served as ZCPs, including GradNorm \cite{abdelfattah2021zero}, Fisher \cite{turner2019blockswap}, SNIP \cite{lee2018snip}, and Synflow \cite{tanaka2020pruning}. \citet{zhou2022training} employed the DSS-indicator to search for improved Vision Transformers (ViT). \citet{serianni2023training} developed a novel proxy attention confidence metric for Transformer-based models, outperforming parameter-only proxies and establishing the FlexiBERT benchmark dataset with 500 architectures and their GLUE performance metrics. To enhance generalization and scoring accuracy, \citet{akhauri2022eznas} implemented genetic programming to automatically discover ZCPs across different search spaces. Similarly, \citet{dong2024lpzero} used genetic search symbolic expressions to find suitable ZCPs for Transformer-based models. While these methods identify better-performing ZCPs, evolutionary search remains computationally expensive, and proxy performance varies with different data inputs. In contrast, our method is data-free, calculating directly from initialization or pre-trained weights. It depends solely on model architecture, eliminating data-related influences.

\subsection{Transformer and Language Models}

Transformer-based language models now underpin many state-of-the-art systems. The Transformer architecture, introduced by \citet{vaswani2017attention}, consists primarily of two key blocks: FFNs and Multi-Head Attention (MHA). MHA enables the model to focus on different input segments across various representation subspaces, enhancing its ability to capture complex data relationships. Subsequently, FFNs introduce non-linearity and provide additional processing to refine the learned representations. Transformers can be designed as encoder-only models (e.g., BERT \cite{devlin2019bert}) utilizing bidirectional attention for tasks like question answering, or as decoder-only models (e.g., GPT-2 \cite{radford2019language}, LLaMA \cite{touvron2023llama}) for text generation by predicting subsequent words in sequences. Throughout this paper, we refer to all MHA-related modules as the Attention block.

\subsection{Model Compression and NAS for LLMs}

Structured pruning effectively reduces computational and storage requirements by removing complete structural units from neural networks while preserving model performance. Several approaches have been applied to LLM pruning, including FLAP \cite{an2024fluctuation}, SliceGPT \cite{slicegpt_iclr24}, Wanda \cite{Sun2023ASA_wanda}, LLM-Pruner \cite{ma2023llm}, Shortened LLaMA \cite{kim2024shortened}, and SLEB \cite{Song2024SLEBSL}. \citet{munoz2024lonas} employed LoRA \cite{hu2022lora} to fine-tune pre-trained weights, creating a supernet with defined search spaces for attention heads and FFNs, enabling the discovery of efficient sub-networks for LLMs. Similarly, \citet{shen2025search} utilized evolutionary search masks and optimization compensation techniques to identify optimal model subnets.

\section{Methodology}\label{sec3}
    \begin{figure*}[ht!]
            \centering
            \includegraphics[width=1\textwidth]{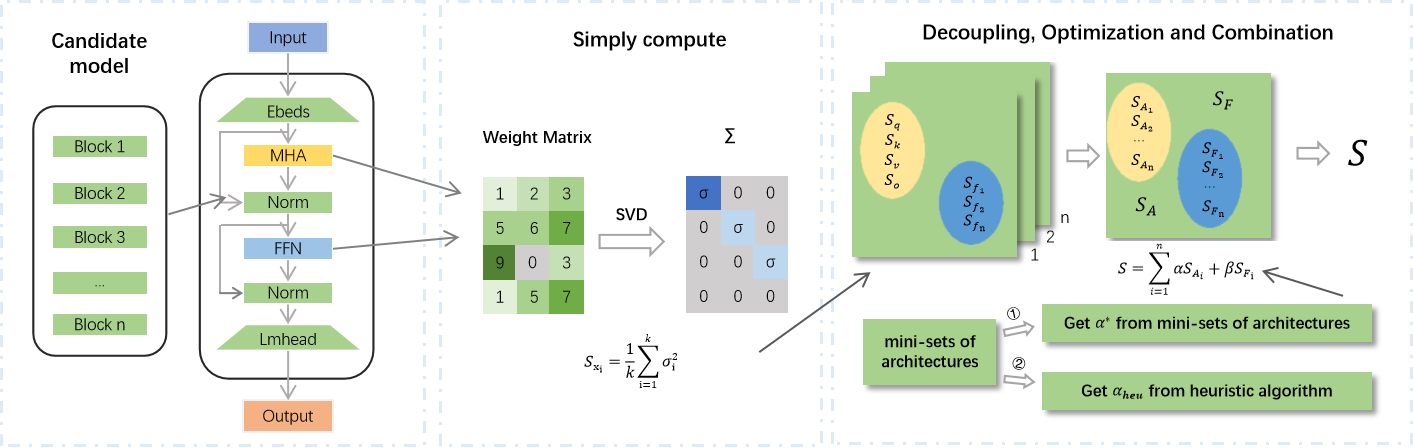}
            \caption{Computing Flow of Proxy Metric.}
            \label{fig2}
    \end{figure*}
\subsection{Theoretical Motivation} 
Our method draws inspiration from advanced theories including generalization theory, complexity analysis, random matrix theory, and deep learning principles. Scaling laws reveal that LLMs' generalization performance exhibits a sublinear relationship with number of parameters \cite{kaplan2020scaling}. Model complexity correlates closely with generalization performance, with number of parameters providing a straightforward complexity measure \cite{belkin2019reconciling}, though this relationship is not strictly monotonic \cite{belkin2019reconciling, ZhangBHRV17, neyshabur2017exploring}. These studies demonstrate that model complexity partially reflects learning capability and generalization performance, which is commonly referred to as model capacity.

Model capacity can be estimated through norms, with numerous studies on weight analysis and spectral analysis establishing connections between model capacity and generalization while estimating generalization error upper bounds \cite{neyshabur2015norm, neyshabur2017exploring}. Building on these theoretical foundations, we propose a straightforward metric to estimate actual model performance from two distinct perspectives.

First, from the model capacity and spectral analysis perspective, we employ singular value decomposition (SVD) to estimate underlying model capacity. Second, from a physical interpretation standpoint, we examine the relationship between matrix energy density and potential learning capacity \cite{saxe2013exact, neyshabur2017exploring}. We hypothesize that untrained model weight matrices should exhibit more evenly distributed energy density, enabling subsequent learning to gradually concentrate representational power into a smaller weight parameter subset, thereby enhancing generalization and learning efficiency \cite{soudry2018implicit}.
     \begin{figure}[ht!]
        \includegraphics[width=0.48\textwidth]{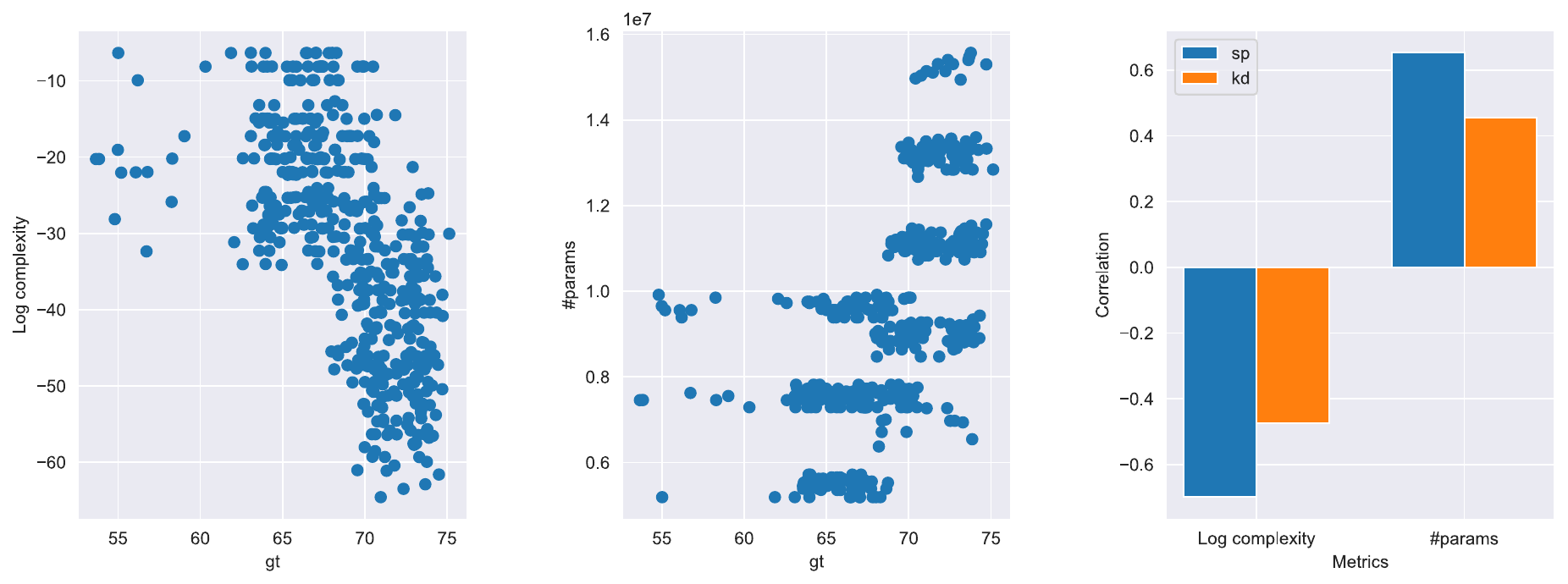}
        \caption{Comparing the performances of Log Complexity and \#params for randomly sampling 500 architectures on the FlexiBERT Benchmark. There is a negative correlation between the Log Complexity and the real performance, and the performance is similar to \#params. This demonstrates the effectiveness of using the norm for measurement, but it is not accurate enough.}
        \label{AFB-log3}
    \end{figure}
This metric inherently relates to model capacity and generalization performance. The Vapnik-Chervonenkis (VC) dimension, which serves as a metric for measuring model capacity, reflects the strength of the model's expressive ability. It correlates with both generalization performance and required training sample count, and can be approximated as $\widetilde{O}(d\times dim(w))$\cite{neyshabur2017exploring}. \citet{martin2021predicting} approximated the VC dimension $\mathcal{C}$ using norm-based methods.
    \begin{equation}
        \mathcal{C} \sim \|\mathbf{W}_1\| \times \|\mathbf{W}_2\| \times \cdots \times \|\mathbf{W}_L\|
    \end{equation}
where $\|\cdot\|$ is a matrix norm, $\mathbf{W}_i$ represents weight matrix of $i$-th network. The log complexity,
    \begin{equation}
        \log \mathcal{C} \sim \log \|\mathbf{W}_1\|+\log \|\mathbf{W}_2\|+\cdots+\log \|\mathbf{W}_L\| 
    \end{equation}
However, when $\sum_{l} \log |\mathbf{W}_l|$ is applied to Transformer-based models, these approaches fail to capture the functional differences between architectural blocks, as demonstrated in Figure~\ref{AFB-log3}, thus reducing their performance. Our method explicitly accounts for these functional variations, providing a more effective proxy for Transformer-based models.
\subsection{The Motivation of Decoupling}
Studies have demonstrated that only a subset of BERT's attention heads contribute meaningfully to downstream tasks \cite{michel2019sixteen}. Both \citet{shen2025search} and \citet{munoz2024lonas} observe that pruning Attention and FFN blocks impacts model performance differently, with Attention blocks showing greater sensitivity to pruning in LLMs. This differential sensitivity reveals that when applying ZCPs to search Transformer-based architectures, significant functional differences exist between Transformer modules. These differences manifest not only in varying NAS search space configurations across model types but also in performance discrepancies stemming from module function differences, affecting the ranking correlation between ZCPs and actual performance. These observations inform our approach: we decouple Attention and FFN blocks to enhance the ranking capability of our ZCP.
\subsection{Proxy Metric}
\citet{zhou2022training} proposed DSS-indicator, which employs distinct evaluation metrics for Attention and FFN blocks. However, this approach maintains dependency on input data. Under certain NAS benchmarks, DSS-indicator underperforms number of parameters, which are robust and relatively stable ZCPs. In contrast, our method utilizes a simple computational metric relying solely on model initialization or pre-trained weights. As shown in Equation~\eqref{eq3}, we perform SVD on all relevant modules within both Attention and FFN blocks. We then extract the singular values, compute the square of each value in the resulting singular value matrix, and average these squared values to obtain the module capacity metric \(S_{xi}\). 
    \begin{equation}\label{eq3}
        S_{x_i}=\frac{1}{\min(n,m)}\sum_{j = 1}^{\min(n,m)}\sigma_{j}^{2}
    \end{equation}
where $n$ and $m$ represent the row and column of the matrix, $\min(n,m)$ represents the length of the shorter side of the matrix, $\sigma_j$ represents $j$-th singular value of matrix.

Essentially, this process is equivalent to calculating the mean squared Frobenius norm of the matrix in Equation~\eqref{eq4}.
    \begin{equation}\label{eq4}
    \begin{array}{ll}
                \|\boldsymbol{W}_i\|_F^2 &= \text{tr}(\boldsymbol{W}_i^T\mathbf{W}_i) = \text{tr}(\boldsymbol{V}\Sigma^T\boldsymbol{U}^T\boldsymbol{U}\Sigma\boldsymbol{V}^T)\\ 
        &=\text{tr}(\Sigma^T\Sigma)=\sum_{i = 1}^{s} \sigma_{i}^{2}
          \end{array}
    \end{equation}
where $\boldsymbol{W}_i$ represents the weight matrix of module $i$ computed, \text{tr($\cdot$)} represents the computation of matrix trace, $\boldsymbol{V}$ represents the left singular vector matrix, $\boldsymbol{U}$ represents the right singular vector matrix, $\boldsymbol{\Sigma}$ represents the singular diagonal matrix.

\noindent\textbf{Balance contribution between attention and FFN}
In order to balance the differences between attention and FFN blocks and their impact on overall model performance, we must account for the varying correlations between proxy scores of these architectural components and actual model performance. Our analysis reveals that attention and FFN contribute to performance outcomes to different degrees across model configurations. To address this variability, we introduce hyperparameter $\alpha$ and $\beta$ that provide precise control over the relative weighting of these components in our evaluation framework.

    \begin{equation}
        S_{attn} = \sum_{mod \in \boldsymbol{A}}{(S_{o},S_{q},S_{v},S_{k},...)}
    \end{equation}
    \begin{equation}
        S_{ffn} = \sum_{mod \in \boldsymbol{F}}{(S_{f1},S_{f2},...)}
    \end{equation}
    \begin{equation}
        S_{proxy} = \sum_{l=1}^{L}{\alpha \times S_{attn}}  + {\beta \times S_{ffn}}
    \end{equation}
where ${\beta = 1 - \alpha}$, $S_{proxy}$ represents the proxy score of architecture, $S_{ffn}$ and $S_{attn}$ represent the proxy score of the FFN block and attention block, $mod$ represents single module, $\boldsymbol{A}$ and $\boldsymbol{F}$ represent modules related to attention and FFN, $S_x$ reprsent proxy score of the module $x$.

    \begin{table*}[ht!]
        \centering
        \resizebox{\textwidth}{!}{
        \begin{tabular}{cccccccccccc}
        \hline
         & \multicolumn{2}{c}{openwebtext(LPZero)} & \multicolumn{2}{c}{wikitext-103} & & \multicolumn{6}{c}{Requirement Comparison} \\
        Proxy & KT & SPR & KT & SPR & Time(s) & Method & Data & Gradient & \#metrics & Computing & Area \\ \hline
        \rowcolor[HTML]{EFEFEF} 
        \#params & 0.454 & 0.652 & 0.454 & 0.652 & - & - & N & N & 1 & - & - \\
        Synflow~\cite{tanaka2020pruning} & 0.322 & 0.471 & -0.320 & -0.470 & 6.805 & manual & A & Y & 1 & backward & P \\
        Fisher~\cite{turner2019blockswap} & 0.139 & 0.209 & -0.148 & -0.224 & 0.539 &manual & A & Y & 1 & backward & P \\
        Gradnorm~\cite{abdelfattah2021zero} & 0.133 & 0.197 & -0.114 & -0.170 & 0.271 &manual & A & Y & 1 & backward & P \\
        SNIP~\cite{lee2018snip} & 0.119 & 0.173 & -0.112 & -0.167 & 0.271 & manual & A & Y & 1 & backward & P \\
        Jacobian cosine~\cite{celotti2020improving} & 0.116 & 0.149 & -0.100 & -0.146 & 0.268 & manual & A & Y & 1 & backward & P \\
        GraSP~\cite{wangpicking} & 0.122 & 0.197 & -0.094 & -0.140 & 1.346 & manual & A & Y & 1 & backward & P \\
        Synaptic Saliency~\cite{tanaka2020pruning} & 0.157 & 0.266 & 0.080 & 0.124 & 0.268 & manual & A & Y & 1 & backward & P \\
        LPZero~\cite{dong2024lpzero} & 0.511 & 0.748 & 0.126 & 0.187 & 2.001 & auto-find & A & Y & 2+ & backward & Z\&P \\
        DSS~\cite{zhou2022training} & - & - & 0.192 & 0.293 & 0.686 & manual & A & Y & 2 & backward & Z \\
        Synaptic Diversity~\cite{zhou2022training} & 0.021 & 0.175 & 0.193 & 0.294 & 0.418 & manual & A & Y & 1 & backward & P \\
        Softmax Confidence~\cite{serianni2023training} & - & - & 0.217 & 0.326 & 0.189 & manual & A & N & 1 & forward & Z \\
        Head importance~\cite{serianni2023training} & 0.050 & 0.171 & 0.221 & 0.331 & 0.272 & manual & A & N & 1 & forward & Z \\
        LogSynflow~\cite{cavagnero2023freerea} & 0.334 & 0.491 & 0.324 & 0.476 &  7.491 & manual & A & Y & 1 & backward & P \\
        EZNAS~\cite{akhauri2022eznas} & 0.483 & 0.698 & 0.479 & 0.696 & 6.680 & auto-find & R,A & Y & 2+ & backward & Z \\
        Attention Confidence~\cite{serianni2023training} & 0.475 & 0.666 & 0.473 & 0.714 & 0.258 & manual & A & N & 1 & forward & Z \\
        Activation Distance~\cite{mellor2021neural} & 0.081 & 0.123 & 0.518 & 0.739 & 0.198 & manual & A & N & 1 & forward & Z \\
        \rowcolor[HTML]{EFEFEF} 
        {\color[HTML]{1D578D} \textbf{Ours}} & {\color[HTML]{1D578D} \textbf{0.532}} & {\color[HTML]{1D578D} \textbf{0.765}} & {\color[HTML]{1D578D} \textbf{0.532}} & {\color[HTML]{1D578D} \textbf{0.765}} & {\color[HTML]{1D578D} \textbf{0.883}} & {\color[HTML]{1D578D} \textbf{manual}} & {\color[HTML]{1D578D} \textbf{N}} & {\color[HTML]{1D578D} \textbf{N}} & {\color[HTML]{1D578D} \textbf{1}} & {\color[HTML]{1D578D} \textbf{-}} & {\color[HTML]{1D578D} \textbf{Z\&P}} \\ \hline
        \end{tabular}}
        \caption{The results of randomly sampling 500 architectures on the FlexiBERT benchmark when the input data is wikitext-103 and openwebtext. The result on openwebtext from LPZero, and the results on wikitext-103 are recalculated, time is average value. Actual data(A), Random noise(R), Yes(Y), No(N), Prunning(P), Zero-shot NAS(Z). }
        \label{ranking flexibert}
    \end{table*}
\subsection{Decouple Optimization Hyperparamters $\alpha$}
The method we provide is straightforward. After decoupling, our optimization objective is to tune the hyperparameter $\alpha$ to enhance our proxy method's ranking ability. We propose two approaches for determining $\alpha$: Firstly, for architectures with established performance benchmarks, we sample a mini-set of real architectures to derive the optimal $\alpha$. Secondly, for search tasks without benchmarks, we offer a heuristic method. By approximating proxy correlations from sampled small-scale data of number of parameters, FFN blocks, and attention blocks, we can determine a relatively optimal $\alpha$ value.
    \begin{algorithm}[ht!]
        \caption{$\alpha^{\star}$ Optimization via Benchmark Sampling}
        \begin{algorithmic}[1]
        \REQUIRE 
        $\boldsymbol{GT} \in \mathbb{R}^n$, $\boldsymbol{ARCH} \in \mathbb{R}^n$\\
        $\alpha$ in the set $\{-1.5, -1.4, \ldots, 1.5\}$\\
        Sample size $k$
        \ENSURE Optimized $\alpha^*$
        
        \medskip
        \STATE \textbf{Step 1: Sampling}
        \STATE Initialize $G \leftarrow \{\}$, $A \leftarrow \{\}$
        \FOR{$i = 1$ to $k$}
            \STATE Randomly select an index $j$ from $\{1,2,\ldots,n\}$
            \STATE Append $\boldsymbol{GT}[j]$ to $\boldsymbol{G}$
            \STATE Append $\boldsymbol{ARCH}[j]$ to $\boldsymbol{A}$
        \ENDFOR
        
        \medskip
        \STATE \textbf{Step 2: Metric Computation}
        \STATE Compute metric vectors $\boldsymbol{S_{F}}$ and $\boldsymbol{S_A}$ from the $\boldsymbol{A}$ 
        \STATE $\boldsymbol{S_F}, \boldsymbol{S_A} \leftarrow \texttt{ComputeMetrics}(\boldsymbol{A})$
        
        \medskip
        \STATE \textbf{Step 3: $\alpha$ Optimization}
        \STATE \textbf{for each} $\alpha$ in $\{-1.5, -1.4, \ldots, 1.5\}$ \textbf{do}
            \STATE \quad $S_{proxy} \leftarrow \texttt{Combinate($\boldsymbol{S_F}$,$\boldsymbol{S_A}$,$\alpha$)}$ 
            \STATE \quad $\tau_{i} \leftarrow \texttt{Kendalltau}(\boldsymbol{G},\boldsymbol{S_{proxy}})$
            \STATE \quad Append $\tau_i$ to $\boldsymbol{T}$
        \STATE \textbf{end for}
        \STATE Get $(\tau_{top-1},\alpha_{top-1}),(\tau_{top-2},\alpha_{top-2}) \leftarrow Sort(T)[0:2]$
        \STATE Compute $\alpha^* \leftarrow (\alpha_{\text{top-1}} + \alpha_{\text{top-2}})/2$
        \RETURN $\alpha^*$
        \end{algorithmic}
        \label{algorithmic:sampling}
    \end{algorithm}
    
These two methods can be summarized as follows.

Sampling correction from the benchmark, for a benchmark dataset with \(n\) architectures, its real performance can be represented as \(\boldsymbol{GT} = [gt_1, gt_2, \ldots, gt_n]\), and its architecture set can be represented as \(\boldsymbol{ARCH} = [A_1, A_2, \ldots, A_n]\). The mini-set we sampled can be denoted as \(\boldsymbol{A} = [A_i, A_j, \ldots, A_k]\), and the corresponding real performance values of the sampled set are \(\boldsymbol{G} = [g_i, g_j, \ldots, g_k]\). The proxy scores we obtained can be represented as \(\boldsymbol{S}_{F} = [S_{f_i}, S_{f_j}, \ldots ,S_{f_k}]\) and \(\boldsymbol{S}_{A} = [S_{a_i}, S_{a_j}, \ldots ,S_{a_k}]\). We calculate KT as the maximization metric to obtain the relatively optimal \(\alpha^{\star}\), The entire optimization process is described as in Algorithm~\ref{algorithmic:sampling}.
    \begin{equation}
        \mathbf{\tau}_{\text{top}-k} = \mathbb{\tau}\left(\boldsymbol{G}, \alpha_{\text{top}-k} \cdot \boldsymbol{S_{A}}+(1 - \alpha_{\text{top}-k})\cdot\boldsymbol{S_{F}}\right)
    \end{equation}
    \begin{equation}
        \alpha^{\star} = \underset{\alpha \in [-1.5, 1.5]}{\max} \frac{1}{2}(\alpha_{\text{top}-1} + \alpha_{\text{top}-2})
    \end{equation}
where $\tau$ represents the calculation of KT between the vectors $\boldsymbol{x}$ and $\boldsymbol{y}$, and $\alpha_{\text{top}-k}$ represents the value of $\alpha$ corresponding to the $k$-th largest $\tau$ value within the current value of $\alpha$.

Another method for optimizing $\alpha$ employs Complete the following expression:
    \begin{equation}
        \resizebox{\columnwidth}{!}{
        $\alpha_{heu} = \underbrace{\big|\max\{\tau_{a,p},\tau_{f,p}\} \times \big(1 - \frac{\tau_{a,f}}{\min\{\tau_{a,p},\tau_{f,p}\}}\big)\big|}_{\text{Dominat-Synergy Adijusted Scaling $\lambda$}} + 
        \texttt{flag}{(\Phi)} \times \underbrace{[(1 - \tau_{f,p}) \times (1 - \tau_{a,p}) + \tau_{a,f}]}_{\text{Joint Contribution Baseline $\Phi$}}$}
    \end{equation}
where $\alpha_{\text{heu}}$ represents the heuristic hyperparameter $\alpha$, and $\tau_{x,y}$ denotes the KT correlation coefficient between vectors $\boldsymbol{x}$ and $\boldsymbol{y}$. Here, $\boldsymbol{a}$, $\boldsymbol{f}$, and $\boldsymbol{p}$ represent the vectors corresponding to the Attn-only component, FFN-only component, and number of parameters of the currently sampled architectures, respectively, while $\texttt{flag}$ represents a sign function.
    \begin{equation}
        \text{$\texttt{flag}$}(\Phi) = 
        \begin{cases}
         1, & \text{if } \Phi > \tau_{a,p} \\
        -1, & \text{if } \Phi \leq \tau_{a,p} 
        \end{cases}
    \end{equation}
and it is determined by the magnitude relationship between the Joint contribution baseline $\Phi$ and $\tau_{a,p}$, which represents the degree of similarity in the contributions between Attn-only and number of parameters.
    \begin{algorithm}[hb!]
        \caption{Heuristic Computation of $\alpha_{\text{heu}}$}
        \begin{algorithmic}[1]
        \REQUIRE Set of $k$ architectures, corresponding $\boldsymbol{A}$ vectors
        \ENSURE Compute Heuristic $\alpha_{heu}$
        \medskip
        
        \STATE \textbf{Step 1: Sampling}
        \STATE Obtain $\boldsymbol{A} \in \mathbb{R}^k$ for $k$ architectures
        \STATE Get \#params $\boldsymbol{P} \in \mathbb{R}^k$ from $\boldsymbol{A}$
        \medskip
        
        \STATE \textbf{Step 2: Metric Computation}
        \STATE Compute $\boldsymbol{S_F}, \boldsymbol{S_A} \leftarrow \texttt{ComputeMetrics($\boldsymbol{A}$)}$
        \medskip
        
        \STATE \textbf{Step 3: $\alpha_{heu}$ Computation}
        \STATE Get $\tau_{a,p},\tau_{f,p},\tau_{a,f} \leftarrow \texttt{Kendalltau($\boldsymbol{x},\boldsymbol{y}$)}$
        \STATE Compute $\Phi \leftarrow \texttt{ComputeJoint{($\boldsymbol{S_A},\boldsymbol{S_F},\boldsymbol{P}$)}}$
        \STATE Sign Discrimination $flag \leftarrow \texttt{flag($\tau_{a,p},\Phi$)}$
        \STATE Compute $\lambda \leftarrow \texttt{ComputeDominate($\boldsymbol{S_A},\boldsymbol{S_F},\boldsymbol{P}$)}$
        \STATE Compute $\alpha_{heu} \leftarrow \texttt{ComputeHeu($\Phi,\lambda,flag$)}$
        \RETURN $\alpha_{heu}$
        \end{algorithmic}
        \label{algorithmic:heu}
    \end{algorithm}
    
The heuristic optimization is described as in Algorithm~\ref{algorithmic:heu}, We explain our heuristic algorithm:
\begin{itemize}
    \item  For the First Term: \textit{Dominant-synergy Adjusted Scaling $\lambda$}
    
This term is designed to identify blocks that correlate more closely with number of parameters. It incorporates corrections by accounting for components that exhibit weak correlations with number of parameters. When the disparity between the two blocks (attention and FFN) relative to number of parameters is excessively large, and the correlation between FFN and attention has a significant gap with the smaller correlation term, this scaling factor is governed by the leading block. Conversely, this term tends toward a relatively small value when correlations are more balanced. The hyperparameter $\alpha$ primarily functions as a mechanism for balancing the attention block contribution. When the absolute value $\alpha$ increases, this term amplifies the contribution of attention to the overall architecture performance evaluation. When FFN is in a dominant position, the value of this term typically remains extremely small. However, this term becomes significantly magnified when attention is the dominant block. This component primarily addresses scenarios where the attention block holds the dominant position in determining architectural performance.
    \item For the Second Term: \textit{Joint Contribution Baseline $\Phi$}

The second term considers the joint coefficient of the inverse correlation between the two blocks (attention and FFN) and number of parameters, along with the correlation terms within the blocks themselves. As blocks become less correlated with number of parameters and more correlated internally, this term increases in magnitude, effectively capturing the performance of the block combination. When the correlation of the combined blocks is weaker than the correlation between attention and number of parameters, it typically indicates that the correlation between the two blocks is feeble while their individual correlations with number of parameters remain robust. This insight motivates the introduction of the sign function $\texttt{flag}$, which helps distinguish between these architectural scenarios.
    
\end{itemize}

\section{Experiments}

In this section, we evaluate our proxy on two primary benchmarks: FlexiBERT and GPT-2. We also assess our proxy on LoNAS-BERT and LoNAS-LLaMA to investigate its efficiency. We compare against baseline ZCPs using SPR and KT correlation coefficients to demonstrate our method's superior ranking ability.

\subsection{Datasets and Settings}
Our aim is to explore the performance and effectiveness of our method across various benchmarks. We conduct different experiments to thoroughly evaluate its capabilities. We utilize the wikitext-103 dataset \cite{merity2017pointer} in the FlexiBERT benchmark established by \citet{serianni2023training}, which includes random sampling from the NAS BERT search space and comprehensive evaluation on GLUE \cite{wang2018glue}. We compare these results with LPZero testing, which uses openwebtext \cite{radford2019language} as mini-batch input data. For the FlexiBERT benchmark, we uniformly set the wikitext-103 mini-batch parameters to a batch size of 128 and token length of 256, matching the configuration used with openwebtext. Similarly, the GPT-2 benchmark established by \citet{javaheripi2022litetransformersearch} includes 200 GPT-2 architectures with test perplexity (PPL) measured on the wikitext-103 dataset. For this benchmark, we set the mini-batch size to 16.

For LoNAS-BERT, we explore ranking capabilities under different actual metrics. We obtain adversarial data samples from TextAttack \cite{morris2020textattack}, a framework integrating mainstream text adversarial methods. We employ PWWs \cite{ren2019generating}, TextBugger \cite{li2019textbugger}, and TextFooler \cite{jin2020bert} as attack methods against the fine-tuned bert-base-uncased model on SST-2 and QNLI tasks. Details of the adversarial data samples are presented in Table~\ref{table:SST-2 and QNLI}.

For the LLaMA search space, we test generative capabilities across eight commonsense tasks: BoolQ \cite{clark2019boolq}, PIQA \cite{bisk2020piqa}, HellaSwag \cite{zellers2019hellaswag}, WinoGrande \cite{sakaguchi2021winogrande}, ARC-Easy and ARC-Challenge \cite{clark2018think}, and SIQA \cite{sap2019socialiqa}. This evaluation demonstrates the effectiveness and efficiency of replacing actual evaluations with our proxy during the search phase. Details of all search spaces are provided in Tables~\ref{tab:LoNAS-Bert search space}, ~\ref{tab:LoNAS-LLaMA search spaces}, ~\ref{table:gpt2 search spaces}(Appendix~\ref{sec7:A}).
    
\subsection{Experiment on FlexiBERT benchmark}
\citet{serianni2023training} constructed a benchmark based on FlexiBERT, which included 500 FlexiBERT models and their actual performance on the GLUE dataset. LPZero \cite{dong2024lpzero} tested the ranking ability of multiple proxy models on the FlexiBERT benchmark using the OpenWebText dataset. Similarly to the experiments in LPZero \cite{dong2024lpzero}, we re-tested these proxy models on the WikiText-103 dataset. We compared the ranking correlation performance of our model between the baseline models tested on WikiText-103 and OpenWebText.
    \begin{figure}[ht!]
            \centering
            \includegraphics[width=0.48\textwidth]{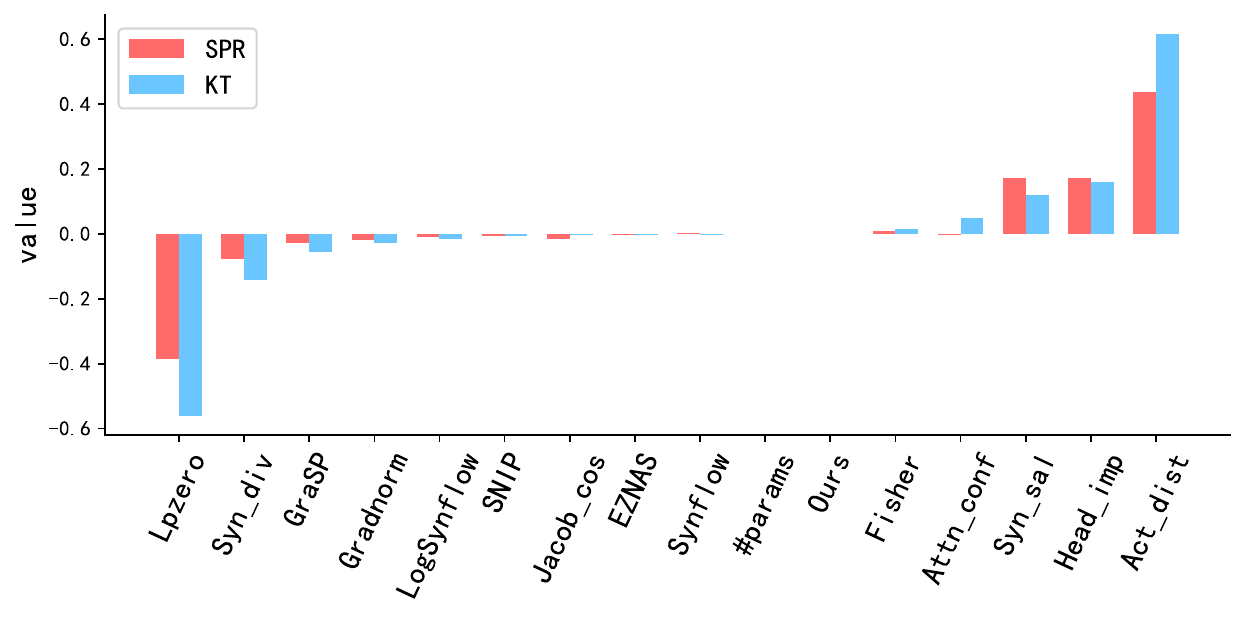}
            \caption{Comparing the changes in the correlation when using openwebtext and wikitext-103 as input data for randomly sampling 500 architectures on the FlexiBERT benchmark.}
            \label{figSPRKT}
    \end{figure}
As shown in Table~\ref{ranking flexibert}, the number of parameters demonstrates strong performance, with most ZCPs unable to exceed the correlation performance of this parameter-based proxy. Several ZCPs exhibit inconsistent performance across the two datasets, particularly head importance, synaptic saliency, synaptic diversity, and activation distance. Activation distance shows the most significant variation, with its KT increasing from 0.081 to 0.518, indicating its high sensitivity to dataset characteristics. Notably, attention confidence consistently outperforms the parameter-based proxy. Figure~\ref{figSPRKT} illustrates how dataset variation affects the ranking capabilities of different ZCPs.

Among automatically discovered proxies, EZNAS demonstrates greater stability. In contrast, the LPZero method, which utilizes the OpenWebText dataset during proxy search, exhibits considerable performance degradation when evaluated with WikiText-103. Our data-free approach achieves KT of 0.533 and SPR of 0.763 on the FlexiBERT benchmark, surpassing all baseline methods and exceeding the parameter-based proxy by margins of 0.079 in KT and 0.113 in SPR. To further illustrate our method's ranking capability, Figure~\ref{fig:plot WT103 flexibert} (Appendix~\ref{sec7:B}) presents a scatter plot comparing GLUE scores against proxy scores on WikiText-103.

We also evaluate the correlation performance of Attn-only and FFN-only when utilized as individual proxies compared to parameter count metrics. Both Attn-only and FFN-only proxies demonstrate superior correlation performance relative to parameter count, with both exhibiting positive correlations. Analysis of the relationships among these four proxies reveals that FFN-only proxies show stronger correlation with parameter count compared to Attn-only proxies. This aligns with the established function of FFN layers in capturing non-linear features, where increased FFN scale generally corresponds to improved performance. In contrast, Attn-only proxies exhibit a weaker correlation with parameter count while maintaining comparable ranking performance, suggesting that the discriminative capability of Attn-only is not predominantly dependent on parameter scale. The observed correlation pattern between Attn-only and FFN-only proxies provides further evidence supporting this conclusion.

    \begin{figure}[ht!]
        \centering
        \includegraphics[width=0.48\textwidth]{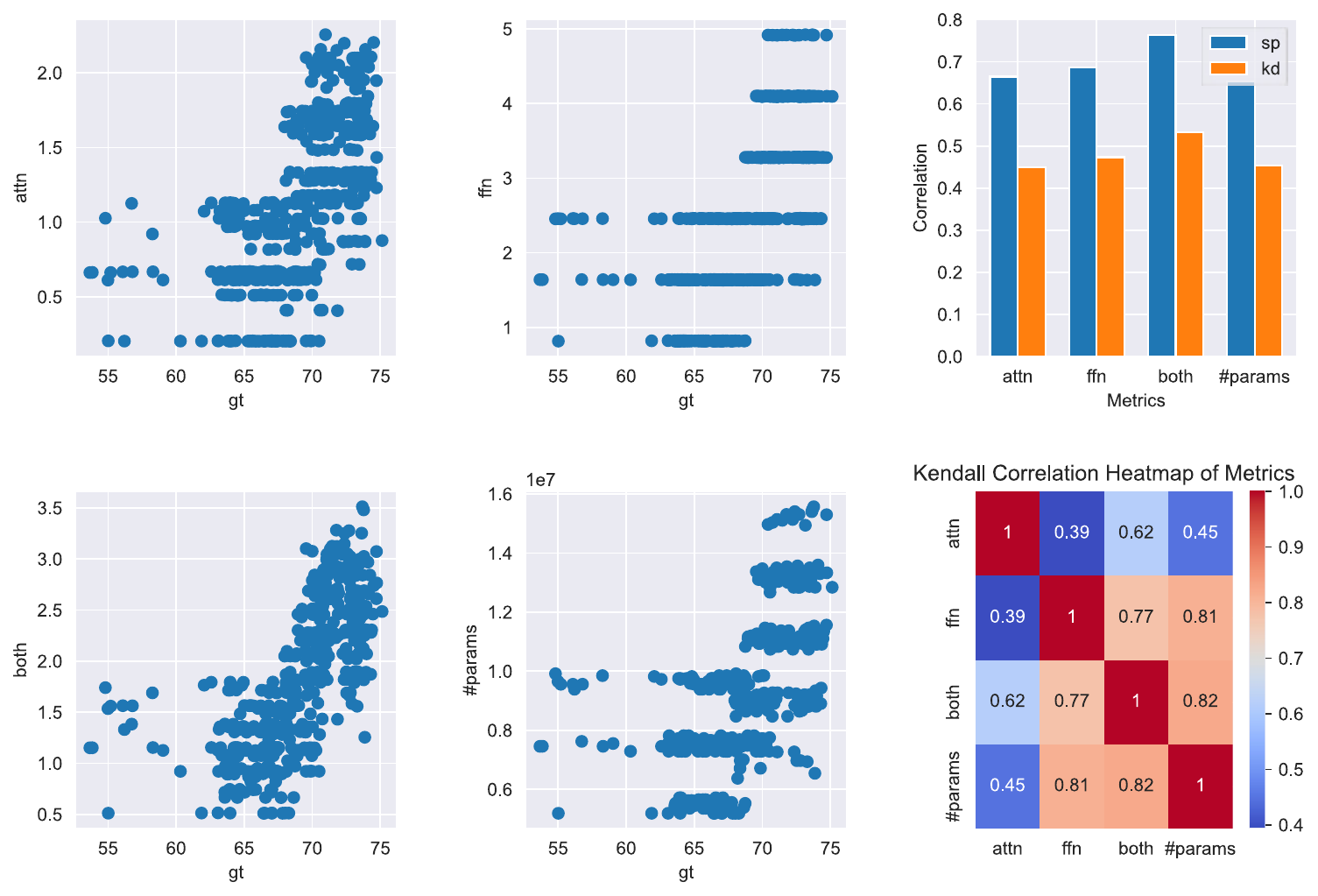}
        \caption{Comparing the performances of FFN-only, Attn-only, Both-only and \#params for randomly sampling 500 architectures on the FlexiBERT Benchmark.}
        \label{figAFB}
    \end{figure}
    
\subsection{Experiment on GPT-2 benchmark}
\citet{javaheripi2022litetransformersearch} randomly samples 200 architectures from the GPT-2 benchmark. Subsequently, we conduct tests on the correlation between these 200 architectures and the test perplexity (PPL) on full wikitext-103. As presented in the Table~\ref{table:gpt-2 ranking}, compared to the \#Decoder.Params~\cite{javaheripi2022litetransformersearch}, and other baseline mothods, our method achieves a strong correlation, with SPR of -0.971 and KT of -0.871. Although our method is marginally less performant than the proxy model acquired through the automated find of LPZero, it remains highly effective.

    \begin{table}[ht!]
        \centering
        \caption{Computation of the proxy's SPR, KT, and time-cost for randomly sampling 200 architectures on the GPT-2 benchmark when the input data is wikitext-103.}
        \begin{tabular}{lccc}
        \hline
        \textbf{Proxy} & \textbf{SPR} & \textbf{KT} & \textbf{time(s)} \\ \hline
        GraSP & 0.483 & 0.380 & 202.9 \\
        EZNAS & -0.602 & -0.439 & 240.3 \\
        Gradnorm & -0.754 & -0.493 & 184.7 \\
        SNIP & -0.779 & -0.506 & 183.7 \\
        Fisher & -0.782 & -0.510 & 336.0 \\
        Jacobian Cosine & -0.735 & -0.532 & 181.5 \\
        Synflow & -0.788 & -0.565 & 299.7 \\
        \#params & -0.737 & -0.582 & - \\
        Activation distance & -0.818 & -0.644 & 180.3 \\
        Attention confidence & -0.851 & -0.677 & 186.9 \\
        Head confidence & -0.851 & -0.678 & 186.1 \\
        LogSynflow & -0.914 & -0.753 & 231.0 \\
        DSS & -0.956 & -0.841 & 494.8 \\
        Synaptic Diversity & -0.956 & -0.841 & 311.9 \\
        \#Decoder.Params & -0.967 & -0.847 & - \\
        Synaptic Synaptic & -0.971 & -0.861 & 183.0 \\
        Head Importance & -0.971 & -0.862 & 186.2 \\
        \textbf{Ours} & \textbf{-0.979} & \textbf{-0.871} & \textbf{260.0} \\
        LPZero & -0.980 & -0.886 & 300.0 \\ \hline
        \end{tabular}
        \label{table:gpt-2 ranking}
    \end{table}
    
\subsection{Experiment on LoNAS}
The experiments on LoNAS involved four metrics. For the experiments based on LoNAS-Bert, we used the clean accuracy(ACC), adversarial accuracy(ADV-ACC), adversarial success rate(ASR), and accuracy drop rate(ACC-Drop) under different adversarial sample datasets as the actual performance of the BERT architecture in text adversarial tasks. For the experiments based on LoNAS-LLaMA, we only used the accuracy of the model on downstream commonsense reasoning tasks as the performance metric.
\subsubsection{LoNAS-Bert for diversity Metrics}
First, we employ the LoNAS framework to sample 1000 architectures derived from the SST-2 task. Subsequently, leveraging the TextAttack toolkits, we initiate attacks on the bert-base-uncased-SST-2 model within the SST-2 dataset and the bert-base-uncased-QNLI model within the QNLI dataset respectively. As a result, three distinct types of adversarial attack samples are acquired. Then, we test the actual performance of these 1000 architectures under diverse adversarial samples to construct a benchmark dataset. The statistics of the sample data are presented in ~\ref{table:SST-2 and QNLI}.

We adopt eight relevant proxies from TRNAS\cite{serianni2023training} as baselines to evaluate the ranking correlation under different real-world performance metrics. In the SST-2 task, we use partial ACC as the real-world data to balance the module differences. Our model calculates the correlation with ACC, ADV-ACC, ASR, and ACC-Drop as the ground truth. Specifically, the correlation remains consistent when using different ACC, ADV-ACC, ASR, and ACC-Drop as the ground truth.
     \begin{figure}[ht!]
        \centering
        \includegraphics[width=0.5\textwidth]{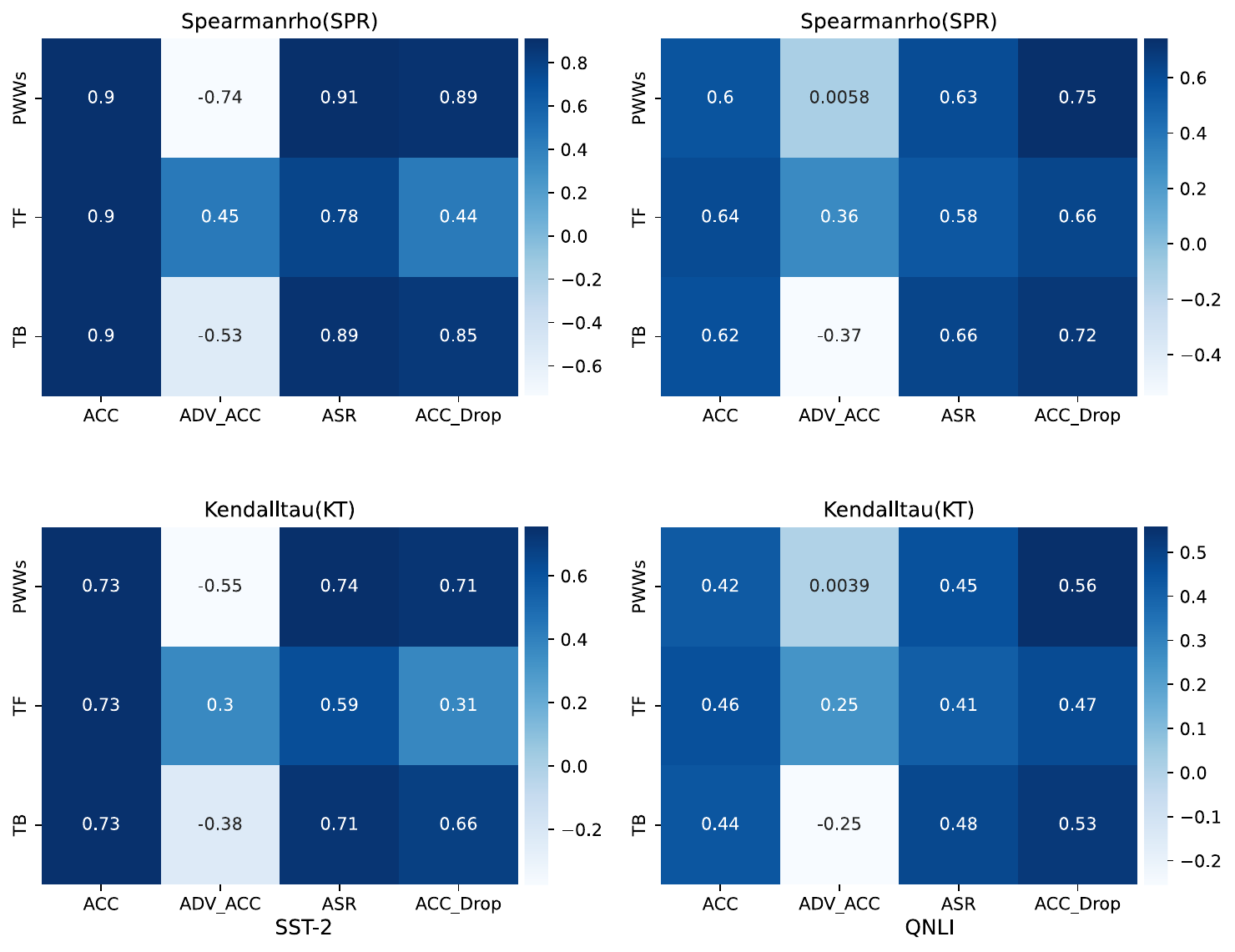}
        \caption{The SPR and KT of 1000 architectures of LoNAS-Bert under 4 different metrics in the evaluation of SST-2 and QNLI adversarial examples for our method against three kinds of attacks}
        \label{bert-heatmap}
    \end{figure}
As results of test on SST-2 show, it diverges for ADV-ACC. It exhibits a negative correlation on PWWs and TextBugger, contrary to the other three metrics, while it aligns on TextFooler. This finding indicates that the proxy model exhibits divergent correlations for attack samples procured via various attack methods with the same metric. Regarding the correlation performance, PWWs and TextBugger show consistency, and the proxy correlation on PWWs is generally stronger than that on TextBugger. The attack samples of TextFooler demonstrate relatively weak correlation. Our model achieves considerably high correlations in all three adversarial samples datasets.

In the context of data transfer performance, we re-evaluate the 1000 architectures on the QNLI adversarial datasets to construct the QNLI architecture benchmark dataset. Similarly, we calculate the correlation performance of the eight baselines and our proxy model. The experimental results show that although our method still maintains a favorable correlation with actual performance, the overall performance experienced a decline compared to the results on SST-2. On the other hand, our method can always achieve some improvement in terms of number of parameters. Although not being the best performing compared to other baselines, our approach still demonstrated some utility and stability across most metrics.

By comparing the test results of our method on two different sample datasets, As shown in the Figure~\ref{bert-heatmap}, we found that there is consistency in the overall trend, while there are inconsistent situations in some individual metrics (such as the tests on PWWs). This may be caused by the differences in the scale of the number of adversarial samples and the deviations of the task objectives. Nevertheless, our method still outperforms most of the other comparative baselines in this test.

\subsubsection{LoNAS-LLaMA for generative abilities and model compression}
Finally, for testing the LoNAS method on the LLaMA-1-7b, we use our proxy to replace the accuracy obtained from the mini-batch real-world evaluation in LoNAS. We then conduct a search in combination with the NSGA-II~\cite{deb2002fast} algorithm. During the search process, we use two objectives: the proxy score to replace the real-world performance and $TFlops$ as an metric to measure the computational complexity of the model. We adopt $TFlops$ as a constraint for the search. Our two objectives are to maximize the proxy score and minimize $TFlops$ in order to find a model with high computational efficiency and good performance. For large-scale and homogeneous models, each module in the search space can be precomputed. We can enables constructing a lookup table to avoid redundant computations during the search phase, significantly the operational cost of the proxy. The objective function is described as follow,
    \begin{equation}
        \begin{aligned}
            \underset{\{ATTN,FFN\}}{{\max}} \quad & (S_{proxy}, -Tflops) \\
            \text{s.t.} \quad &  \text{$Low Params$} < P < \text{$High Params$}
        \end{aligned}
    \end{equation}
where $S_{proxy}$ represent proxy score, $Tflops$ represent Tera floating point operations per second, $P$ represent number of currently architecture paramters, $\{ATTN,FFN\}$ represents the architectures that can be searched within the current search space.

We compared our method with heuristic subnet and subnet in terms of generative ability to verify the effectiveness of our proxy. Our method demonstrated high efficiency and a certain degree of effectiveness in the Figure~\ref{tab:generative ability}. The models searched within 25 minutes had an average accuracy higher than that of the heuristic subnet and close to the performance of the evolutionary search subnet.

Similar to LPzero, we also compared the performance of the structured pruning model with that of our proxy model. As the Table~\ref{table10:Purning method for 8 commensence task} shows, when compared with the structured pruning model, our method showed better performance in commonsense reasoning tasks than other structured pruning methods.

    \begin{table*}[ht!]
        \centering
        \caption{We compared the heuristic subnet, the subnet obtained by evolutionary search for 36 hours using the real evaluation metric, and the two groups of subnets obtained by evolutionary search under different parameters using our ZCP as the metric, and compared the accuracy performance on 8 common sense reasoning tasks.}
        \begin{adjustbox}{width=\textwidth}
        \begin{tabular}{ccccccccccccc}
        \hline
         & ARC-C & ARC-E & Boolq & Hellaswag & OBQA & PIQA & SIQA & Winogrande & Avg & \#params & Tflops & Times(min) \\ \hline
        LoNAS-Heuristics & \textbf{0.585} & 0.723 & 0.623 & 0.513 & 0.710 & 0.729 & 0.658 & 0.639 & 0.647 & 5.5B & 1.44 & - \\
        LoNAS-Subnet(36hour) & 0.581 & \textbf{0.724} & \textbf{0.646} & \textbf{0.524} & 0.724 & \textbf{0.738} & 0.686 & \textbf{0.643} & \textbf{0.658} & 5.6B & 1.43 & 2160 \\
        LoNAS+Ours1 & 0.535 & 0.715 & 0.610 & 0.508 & \textbf{0.735} & 0.690 & 0.639 & 0.639 & 0.634 & 5.6B & 1.45 & 20+5 \\
        LoNAS+Ours2 & {\textbf{0.585}} & 0.714 & 0.640 & 0.513 & 0.722 & 0.732 & \textbf{0.691} & 0.633 & 0.654 & 5.4B & 1.40 & 20+5 \\ \hline
        \end{tabular}
        \end{adjustbox}
        \label{tab:generative ability}
    \end{table*}

\section{Ablation Study}\label{sec4}
\begin{figure}[hb!]
        \centering
        \includegraphics[width=0.48\textwidth]{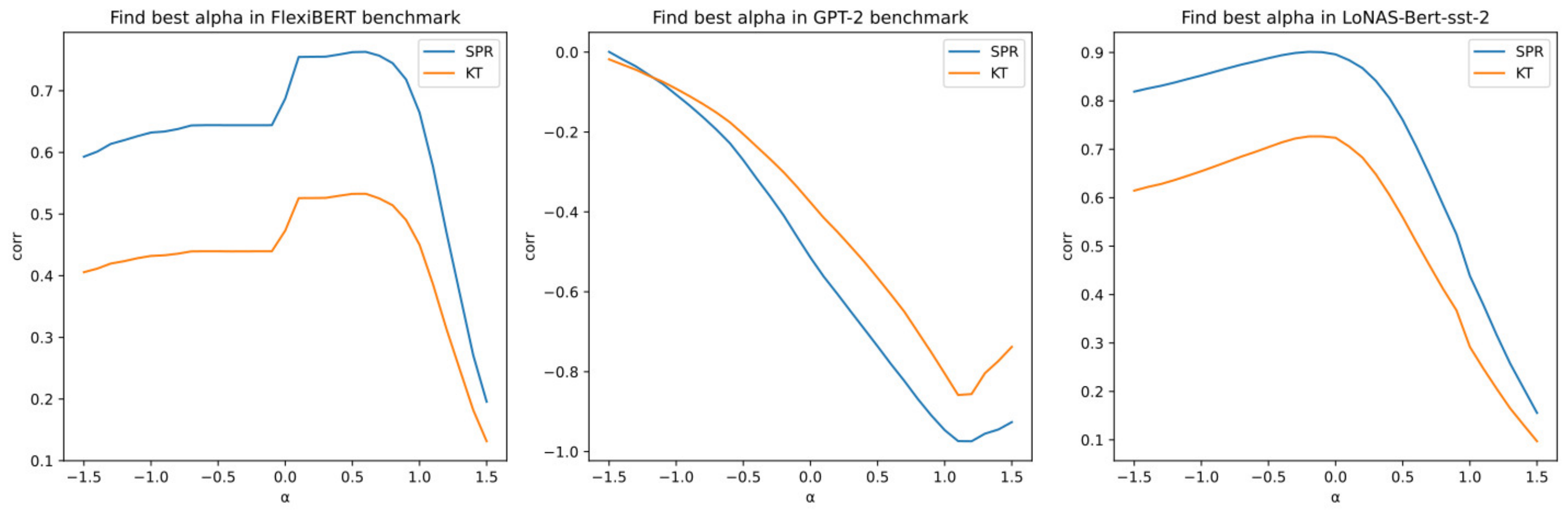}
        \caption{Using the method of benchmark sampling to optimize the hyperparameter $\alpha$ in the FlexiBERT, GPT-2, and LoNAS-Bert benchmarks describes the curves of how SPR and KT change with $\alpha$ in the range from \(-1.5\) to \(1.5\).}
        \label{figabt}
    \end{figure}

\begin{figure*}[ht!]
        \centering
        \includegraphics[width=1\linewidth]{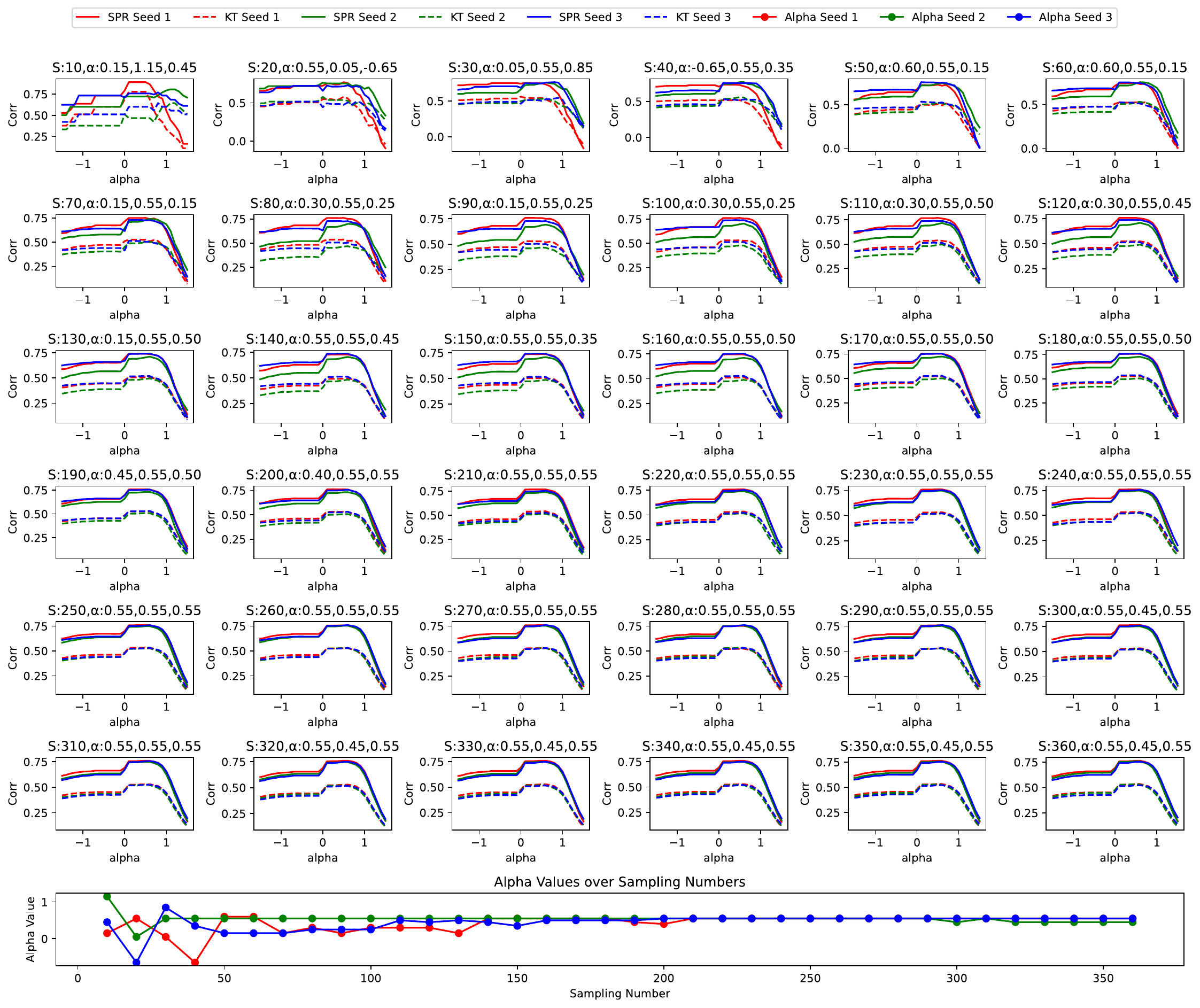}
        \caption{In the FlexiBERT benchmark, when using benchmark sampling direct optimization with the $\alpha$ optimization range set from -1.5 to 1.5, the variations in SPR and KT with changes in $\alpha$, the number of samples, and seeds are examined, along with the convergence behavior of the optimal $\alpha$ under different seeds and sample sizes.}
        \label{fig:alpha-corr-seed3}
    \end{figure*}

\noindent\textbf{Hyperparameter $\mathbf{\alpha}$. }
As shown in the Figure~\ref{fig:alpha-corr-seed3}, in the FlexiBERT benchmark, SPR and KT gradually tend to stabilize as the number of samples increases, and the impact of random seeds on performance is significantly reduced with more samples. For $\alpha$, as the number of samples increase, the optimal alpha parameter gradually converges to fluctuate within a narrow range. Combined with the optimization process plots under different sample sizes, it can be observed that within the optimal interval, there is no significant difference in the SPR and KT achieved by $\alpha$ values on either side. This indicates that even with a mini-batch samples, a high-performing $\alpha$ can directly optimized in a single run.As shown in the Figure~\ref{figabt}, $\alpha$ on benchmark, the optimizable range of the hyperparameter $\alpha$ is relatively narrow. In constrast, on the LoNAS-Bert benchmark, the optimizable range of $\alpha$ is smaller than that on FlexiBERT but larger than that on GPT-2, this indicates that different benchmark. This implies that under different search spaces, the optimization sensitivity of our method varies. These results also demonstrate that our approach of not directly adopting the optimal $\alpha$ obtained from the current sampling is effective in preventing us from falling into a local optimum. Moreover, as the number of samplings increases, this method of optimizing $\alpha$ becomes more stable.
    \begin{figure*}[htb!]
        \centering
        \includegraphics[width=1\textwidth]{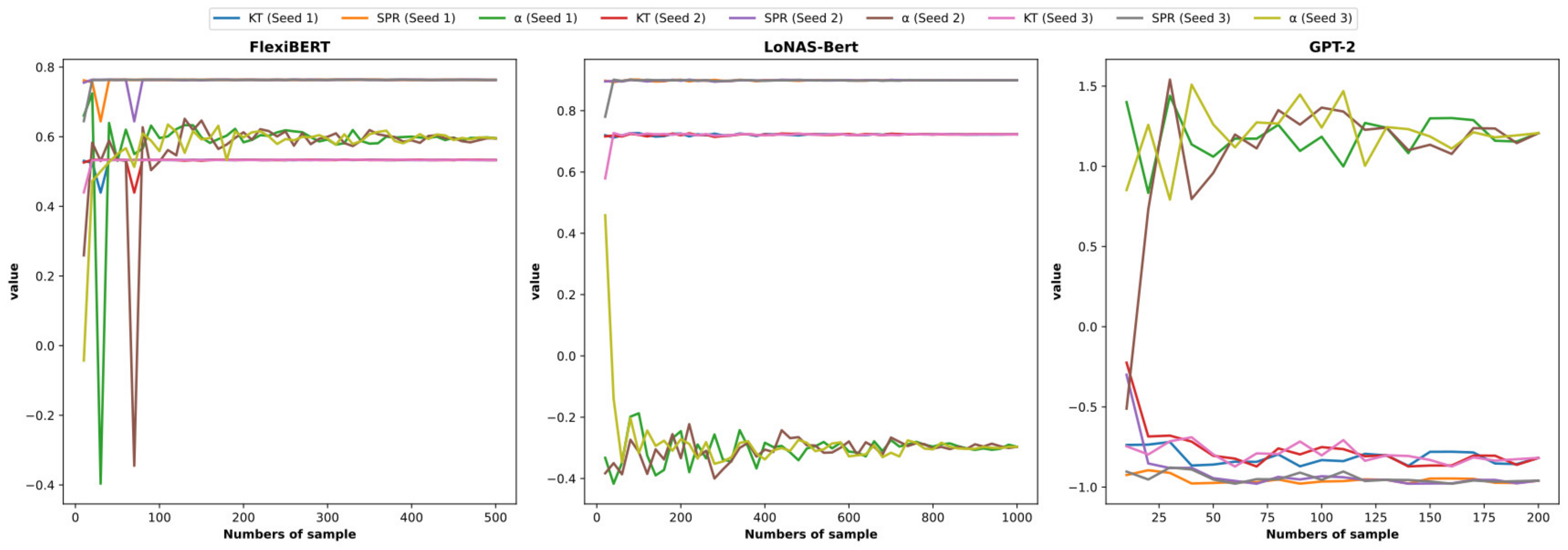}
        \caption{Using the heuristic calculation method to optimize the hyperparameter $\alpha$ in the FlexiBERT, GPT-2, and LoNAS-Bert benchmark, and observe the changes of $\alpha$, SPR, and KT with the variation of the number of samples and random seeds.}
        \label{figabh}
    \end{figure*}
    
\noindent\textbf{Heuristic parameters $\alpha_{heu}$ and numbers of samples. }
As shown in the Figure~\ref{figabh}, the results obtained by optimizing $\alpha$ using the heuristic algorithm indicate that, by observing the changes with different seeds and the number of samples on various benchmarks, the $\alpha$ determined by the heuristic method is effective. The $\alpha$ values under different random seeds on FlexiBERT are unstable when the number of samples is small, while they are relatively stable on GPT-2 and LoNAS-Bert. In terms of correlation, the $\alpha$ values on FlexiBERT and GPT-2 are very stable, and the relative stability on GPT-2 may be due to the fact that the benchmark samples used for testing are not large enough. These results demonstrate the effectiveness and stability of our heuristic algorithm in optimizing the $\alpha$.
    \begin{figure*}[htb!]
            \centering
            \includegraphics[width=1\textwidth]{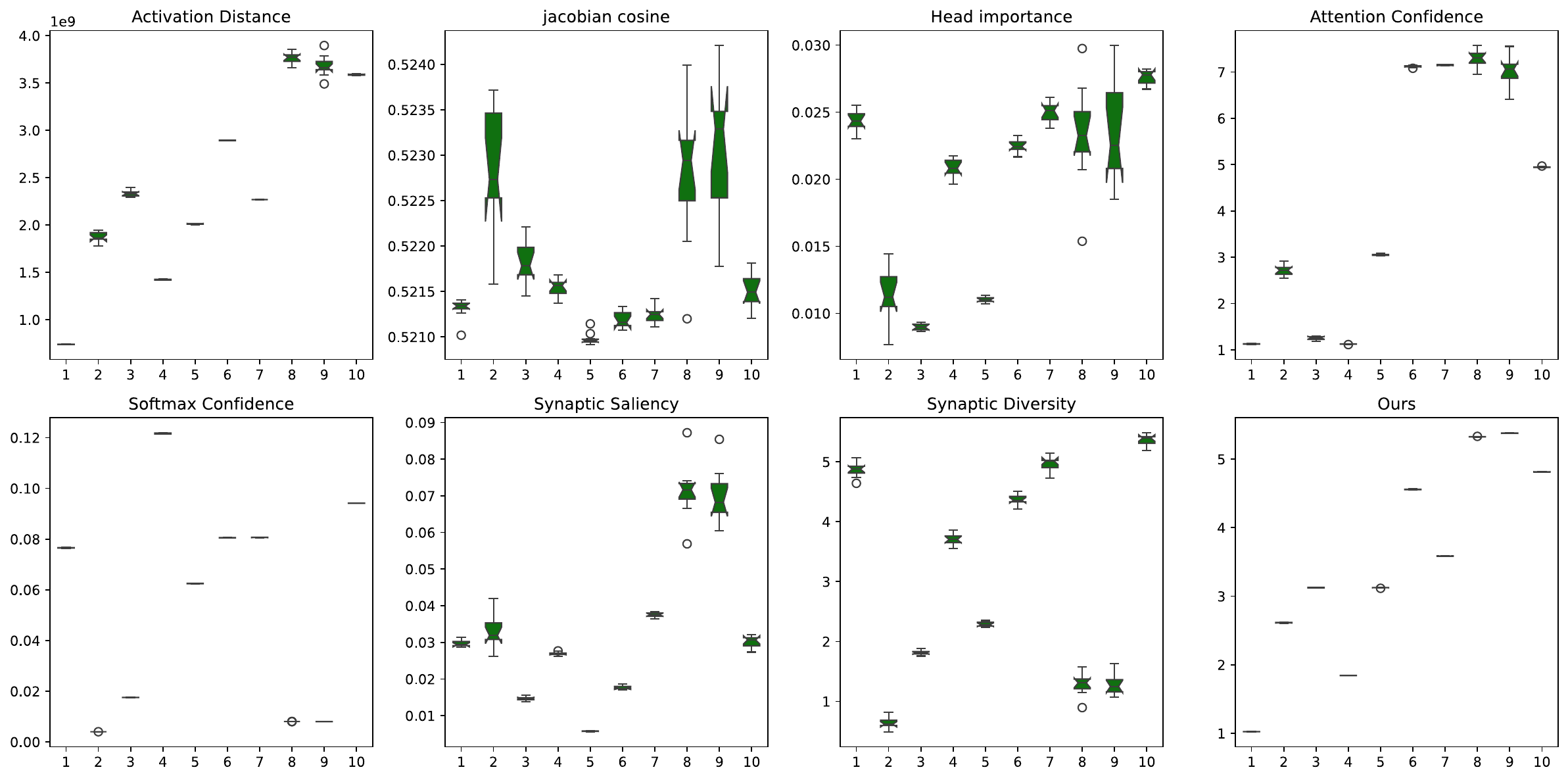}
            \caption{Evaluation of zero-shot metrics with various initialization weights in the FlexiBERT search space. Ten architectures are randomly sampled from the search space, representing decile ranges of the GLUE score (e.g., 0-10\%, 10-20\%, ..., 90-100\%). To ensure robustness, ten different random seeds are employed for weight initialization.}
            \label{fig:random weigth zc}
    \end{figure*}
    \begin{figure*}[htb!]
            \centering
            \includegraphics[width=1\textwidth]{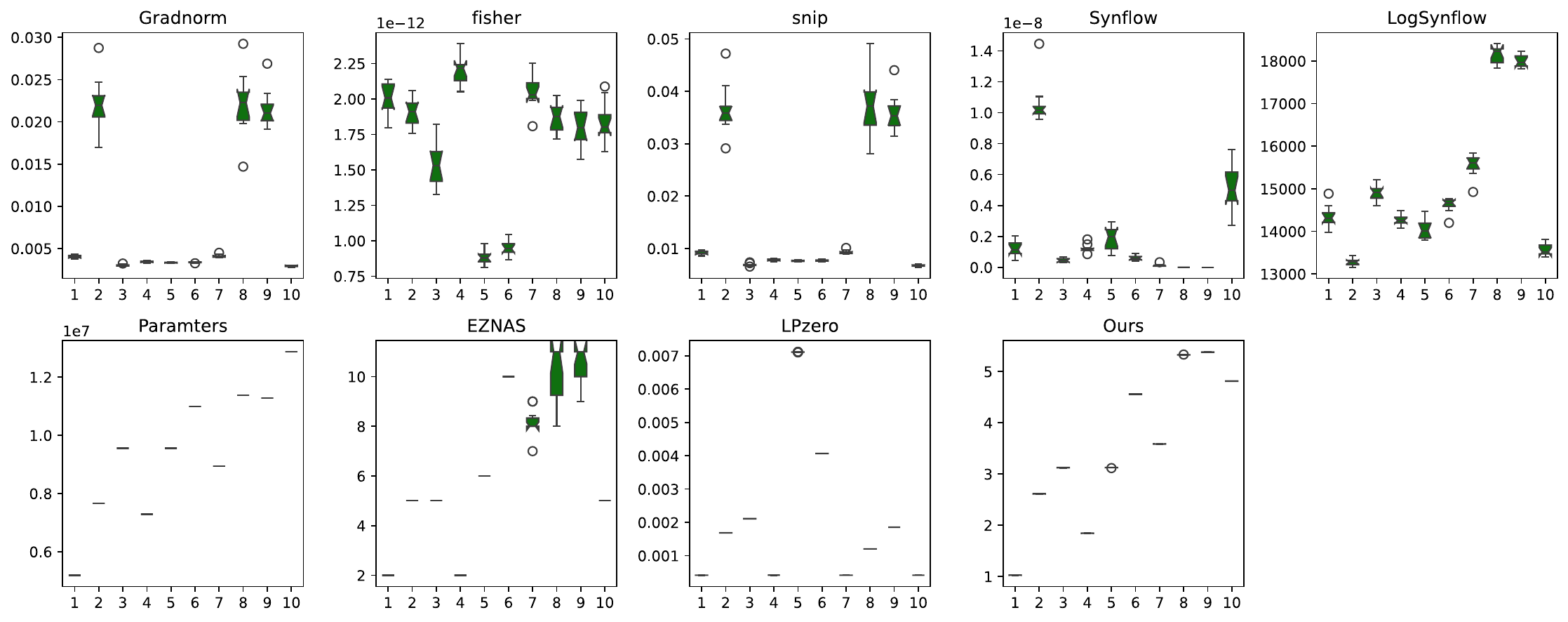}
            \caption{Evaluation of Pruning metrics as ZCP from automatic find with various initialization weights in the FlexiBERT search space. Ten architectures are randomly sampled from the search space, representing decile ranges of the GLUE score (e.g., 0-10\%, 10-20\%, ..., 90-100\%). To ensure robustness, ten different random seeds are employed for weight initialization.}
            \label{fig:random weigth pruning}
    \end{figure*}

\noindent\textbf{Random seed and weight initalization }

To assess the stability of zero-shot metrics, we investigated the impact of random architecture initialization on metric evaluation within the FlexiBERT search space. As illustrated in Figure~\ref{fig:random weigth zc}, several ZCPs demonstrate significant instability, with Jacobian cosine and head importance metrics showing pronounced performance fluctuations across multiple test intervals. Other metrics exhibit relative stability, with softmax confidence and our proposed method remaining virtually unaffected by different random seeds. Comparative analysis between pruning metrics, parameter count proxies, and automatically discovered ZCPs in Figure~\ref{fig:random weigth pruning} reveals that most metrics display localized instability, with only parameter count, LPZero, and our method maintaining consistent performance. While our approach operates on randomly initialized matrices, its correlation performance remains independent of initialization randomness. These results provide compelling evidence for the robust stability of our proposed method across varying initialization conditions.



    \begin{table*}[t]
        \centering
        \caption{A comparative analysis based on LLaMA-1-7b as the pruned model, we compared the performance of several different structured pruning models and an evolutionary search LLM NAS method on 7 tasks.}
        \begin{adjustbox}{width=\textwidth}
        \begin{tabular}{cccccccccc}
        \hline
        \textbf{Pruning method} & \textbf{BoolQ} & \textbf{PIQA} & \textbf{HellaSwag} & \textbf{WinoGrande} & \textbf{Arc-e} & \textbf{Arc-c} & \textbf{QBQA} & \textbf{Avg} & \textbf{\#params} \\ \hline
        LLaMA-7b & 73.18 & 78.35 & 72.99 & 67.01 & 67.45 & 41.38 & 42.40 & 63.25 & 6.7B \\
        SliceGPT~\cite{slicegpt_iclr24} & - & 66.87 & 63.38 & 54.16 & 58.46 & 34.56 & \textbf{-} & 55.49 & 5.0B \\
        Wanda(LoRA)~\cite{Sun2023ASA_wanda} & 72.90 & - & 55.36 & 67.48 & 71.42 & 37.97 & 29.80 & 55.82 & - \\
        SLEB~\cite{Song2024SLEBSL} & {\color[HTML]{333333} \textbf{-}} & 73.07 & {\color[HTML]{34696D} 58.96} & 62.47 & 56.48 & 33.02 & \textbf{-} & 56.80 & 5.5B \\
        LLM-Pruner~\cite{ma2023llm} & 59.39 & 75.57 & 65.34 & 61.33 & 59.18 & 37.12 & 39.80 & 56.82 & 5.4B \\
        Wanda(Full)~\cite{Sun2023ASA_wanda} & 74.50 & - & 55.83 & 69.02 & 73.49 & 39.20 & 32.20 & 57.37 & - \\
        LLM-Pruner(LoRA)~\cite{ma2023llm} & 69.54 & 76.44 & 68.11 & 65.11 & 63.43 & 37.88 & 40.00 & 60.07 & 5.4B \\
        Shortened LLaMA~\cite{kim2024shortened} & 72.70 & 75.70 & 70.40 & 63.60 & 69.50 & 30.10 & 41.20 & 61.89 & 5.5B \\
        FLAP~\cite{an2024fluctuation} & 69.63 & 76.82 & 71.20 & 68.35 & 69.91 & 39.25 & 39.40 & 62.08 & 5.5B \\
        Search for ELLM~\cite{shen2025search} & 74.37 & 76.88 & 70.71 & 67.56 & 68.39 & 40.10 & 39.20 & 62.46 & 6.0B \\
        Ours & 64.00 & 73.20 & 51.30 & 63.30 & \textbf{71.40} & \textbf{58.50} & \textbf{72.20} & \textbf{64.84} & 5.4B \\ \hline
        \end{tabular}
        \end{adjustbox}
        \label{table10:Purning method for 8 commensence task}
    \end{table*}
\section{Discussion}\label{sec5}
\subsection{Zero-shot NAS and NAS for Supernet Pruning} 
The models provided by zero-shot NAS are fully initialized and independent of training processes. In contrast, NAS for Pruning employs NAS-based structural pruning within a supernet with pre-trained weights, which may exhibit spectral distributions different from those of initialized weights. The impact on model capacity is not directly reflected in the training process but is rather determined by the model architecture itself. For NAS for Pruning, the optimization interval of $\alpha$ demonstrated by our method has a smaller impact on ranking ability, while other ZCPs also show improved performance. These results may be attributable to the training process. Our method validates its effectiveness across both types of NAS frameworks by decoupling the Transformer components and optimizing the parameters systematically. Based on the overall outcomes, our method appears more effective for initialized zero-shot NAS compared to pruning-based approaches. When examining baseline methods, we observe that the performance proxy transmitted by the pre-trained supernet for mini-batch inputs likely provides a closer approximation to real-world performance than that of merely initialized models.

\subsection{Encoder-only and Decoder-only} 
The test results on the FlexiBERT and GPT-2 benchmarks demonstrate that our method performs effectively across both Encoder-only and Decoder-only architectures, with no significant performance disparity between them. When optimizing the $\alpha$, we observe that the optimal ranges for $\alpha$ differ between these two benchmarks. This suggests that the determining conditions for optimal $\alpha$ values may not solely depend on the architectural paradigm (encoder versus decoder) but rather on the specific characteristics of the search space. Our findings indicate that both of our proposed optimization methods provide relatively stable hyperparameter optimization effects, consistently enhancing the ranking ability across different transformer architectures and search spaces.

\subsection{Limitations.} 
We evaluated our model across two zero-shot NAS benchmarks and one NAS-for-pruning framework. The proposed approach demonstrated exceptional ranking capabilities and shows promise as an effective pruning metric for Transformer-based architectures. Hardware constraints limited both the diversity of test data across different benchmarks and the number of architectures in our evaluation. Due to these computational resource limitations, validation on larger benchmark datasets remains for future work.

\section{Conclusion}\label{sec6}
We presented a simple, stable, data-free and training-free metric for Transformer-based architecture search. Our approach decoupled the Transformer into two functional components and optimized the hyperparameter $\alpha$ through an efficient single-step process, enabling seamless integration into existing Transformer-based model search frameworks. We introduced two optimization strategies: benchmark-based sampling optimization and a self-parameter correlation heuristic method that operates without requiring benchmark datasets. Empirical evaluation across three distinct NAS frameworks demonstrated the robust performance of our approach. The proposed method provides a powerful yet straightforward metric for Transformer-based model search and can be effectively applied to language model pruning tasks.

{
\bibliographystyle{IEEEtranN}
\bibliography{./main.bib}
}

\newpage
\appendix\label{sec7}

\subsection{Search Space Details}\label{sec7:A}
    This section provides a detailed description of the search space for the NAS method used.The search space of LoNAS-Bert in Table~\ref{tab:LoNAS-Bert search space}, the search space of LoNAS-LLaMA in Table~\ref{tab:LoNAS-LLaMA search spaces}, These search spaces are based on a super-network and are homogeneous, with similarities to structured pruning. The search space of GPT-2 in Table~\ref{table:gpt2 search spaces}. The search space of FlexiBERT in Table~\ref{flexibert search space} is heterogeneous.
    \begin{table*}[h!]
        \centering
        \caption{Search spaces for the BERT-base super-network. some different adjustments have been made to the Attention dimensions and FFN dimensions of each layer. For the LoRA-Rank, the same search space is adopted for each layer. The total search space size of $3^{25} \times 2^{11}$.}
        \begin{tabular}{cccc}
        \hline
        \textbf{Layer} & \textbf{Q\&K\&V\&Q-LoRA\&V-LoRA} & \textbf{LoRA-Rank} & \textbf{Intermediate Dense} \\ \hline
        0 & {[}768, 384{]} & {[}8, 4, 2{]} & {[}3072, 2634, 216{]} \\
        1 & {[}768, 320{]} & {[}8, 4, 2{]} & {[}3072, 2634, 181{]} \\
        2 & {[}768, 256{]} & {[}8, 4, 2{]} & {[}3072, 2627, 208{]} \\
        3 & {[}768, 512{]} & {[}8, 4, 2{]} & {[}3072, 2676, 226{]} \\
        4 & {[}768, 512{]} & {[}8, 4, 2{]} & {[}3072, 2628, 179{]} \\
        5 & {[}768, 704{]} & {[}8, 4, 2{]} & {[}3072, 2662, 175{]} \\
        6 & {[}768, 576{]} & {[}8, 4, 2{]} & {[}3072, 2706, 182{]} \\
        7 & {[}768, 576{]} & {[}8, 4, 2{]} & {[}3072, 2687, 169{]} \\
        8 & {[}768, 640{]} & {[}8, 4, 2{]} & {[}3072, 2616, 165{]} \\
        9 & {[}768, 192{]} & {[}8, 4, 2{]} & {[}3072, 2400, 160{]} \\
        10 & {[}768, 704, 192{]} & {[}8, 4, 2{]} & {[}3072, 2198, 163{]} \\
        11 & {[}768, 320{]} & {[}8, 4, 2{]} & {[}3072, 1940, 150{]} \\ \hline
        \end{tabular}
        \label{tab:LoNAS-Bert search space}
    \end{table*}
    
    \begin{table*}[h!]
        \centering
        \caption{Search spaces for the LLaMA-1-7B super-network, for each layer, the same search space is adopted. In this space, the Q, K, and V of each layer in the attention module are all enhanced by LoRA-rank. The dimension design of the Up and Gate matrices includes five different widths, and the Up and Gate are also enhanced by LoRA-rank. The total search space size of $2^{64} \times 5^{32}$}
        \begin{tabular}{cccc}
        \hline
        \textbf{Model} & \textbf{LoRA-Rank(QKV)} & \textbf{\begin{tabular}[c]{@{}c@{}}Up \& Gate\\ \& Up-LoRA,Gate-LoRA\end{tabular}} & \textbf{LoRA-Rank(Up, Gate)} \\ \hline
        \multicolumn{1}{r}{LLaMA-1-7B} & {[}32, 28{]} & \multicolumn{1}{r}{{[}11008, 9632, 8256, 6880, 5504{]}} & {[}32, 28{]} \\ \hline
        \end{tabular}
        \label{tab:LoNAS-LLaMA search spaces}
    \end{table*}

    \begin{table*}[h!]
        \centering
        \caption{The GPT-2 benchmark, covering a broad spectrum of architectural configurations. Once a model dimension (dmodel) is chosen, the minimum inner dimension (dinner) is set to twice the value of dmodel to avoid training collapse. This adaptive approach ensures a wide range of effective and efficient architectures, summing up to more than $10^{54}$ unique configurations.}
        \begin{tabular}{lc}
        \hline 
        \textbf{Architecture Element} & \textbf{Hyperparameters Values} \\
        \hline 
        Number of Layers (nlayer) & $\{2, 3, ..., 16\}$ \\
        Model Dimension (dmodel) & $\{128, 192, ..., 1024\}$ \\
        Inner Dimension (dinner) & $\{256, 320, ..., 4096\}$ \\
        Number of Attention Heads (nhead) & $\{2, 4, 8\}$ \\
        Adaptive Input Embedding Dimension (dembed) & $\{128, 256, 512\}$ \\
        Adaptive Input Embedding Factor (k) & $\{1, 2, 4\}$ \\
        \hline
        \end{tabular}
        \label{table:gpt2 search spaces}
    \end{table*}

    \begin{table*}[hbt!]
        \centering 
        \caption{The FlexiBERT search space, with hyperparameter values spanning those found in BERT-Tiny and BERT-Mini. Hidden dimension and number of encoder layers is fixed across the whole architecture; all other parameters are heterogeneous across encoder layers. The search space encompasses 10,621,440 architectures.}
        \begin{tabular}{ll}
        \hline
        \textbf{Architecture Element} & \textbf{Hyperparameters Values} \\
        \hline
        Hidden dimension & \{128, 256\} \\
        Number of Encoder Layers & \{2, 4\} \\
        Type of attention operator & \{self-attention, linear transform, span-based dynamic convolution\} \\
        Number of operation heads & \{2, 4\} \\
        Feed-forward dimension & \{512, 1024\} \\
        Number of feed-forward stacks & \{1, 3\} \\
        Attention operation parameters & \\
        \quad if self-attention & \{scaled dot-product, multiplicative\} \\
        \quad if linear transform & \{discrete Fourier, discrete cosine\} \\
        \quad if dynamic convolution & \ convolution kernel size: \{5, 9\} \\
        \hline
        \end{tabular}
         \label{flexibert search space}
    \end{table*}
    
\subsection{Adversarial Examples sets Details}\label{sec7:B}
    The details of adversarial attack samples are described in Table~\ref{table:SST-2 and QNLI}, SST-2 is a single-sentence classification task that contains sentences from movie reviews and their human-annotated sentiment labels. We get adversarial attack samples by attack Bert-base-uncased turning in SST-2. QNLI is a double-sentences classification task that contains sentences from SQuAD, labels is entailment and non-entailment, aim to test performance of models in question answering problems.

    Table~\ref{table:PWWs-SST2}, Table~\ref{table:TB-SST2} and Table~\ref{table:TF-SST2}, SPR and KT of other ZCPs under different attacks and metrics for the LoNAS-Bert-SST2 and -QNLI task. Most metrics exhibit poor performance, while a few show moderate effectiveness in specific scenarios.
    
     \begin{table*}[ht!]
        \centering
        \caption{Statistics of adversarial attack samples of SST-2 and QNLI against bert-base-uncased on PWWs, TextBugger (TB), and TextFooler (TF) obtained from the TextAttack tool.}
        \begin{tabular}{ccccccccccccl}
        \hline
        \textbf{Datasets} & \multicolumn{6}{c}{\textbf{SST2}} & \multicolumn{6}{c}{\textbf{QNLI}} \\ 
        \textbf{Attacker} & \multicolumn{2}{c}{PWWs} & \multicolumn{2}{c}{TF} & \multicolumn{2}{c}{TB} & \multicolumn{2}{c}{PWWs} & \multicolumn{2}{c}{TF} & \multicolumn{2}{c}{TB} \\ \hline
        Label & 0 & 1 & 0 & 1 & 0 & 1 & 0 & 1 & 0 & 1 & 0 & 1 \\
        Nums & 319 & 379 & 236 & 319 & 359 & 408 & 2121 & 2406 & 2479 & 2324 & 2482 & 2162 \\
        \#Query & \multicolumn{2}{c}{{\color[HTML]{333333} 140.411}} & \multicolumn{2}{c}{{\color[HTML]{34696D} 41.377}} & \multicolumn{2}{c}{86.28} & \multicolumn{2}{c}{220.323} & \multicolumn{2}{c}{65.464} & \multicolumn{2}{c}{110.607} \\
        \#preturb & \multicolumn{2}{c}{0.841} & \multicolumn{2}{c}{0.772} & \multicolumn{2}{c}{0.781} & \multicolumn{2}{c}{0.786} & \multicolumn{2}{c}{0.725} & \multicolumn{2}{c}{0.703} \\ \hline
        \end{tabular}
        \label{table:SST-2 and QNLI}
    \end{table*}

    \begin{table*}[ht!]
        \centering
        \caption{The four metrics tested on the SST-2 and QNLI adversarial samples obtained using the TF method are used as the SPR and KT between the performance and 9 different proxies.}
        \label{table:TF-SST2}
        \resizebox{\textwidth}{!}{
        \begin{tabular}{ccccccccccccccccc}
        \hline
        \textbf{Datasets} & \multicolumn{8}{c}{\textbf{SST-2}} & \multicolumn{8}{c}{\textbf{QNLI}} \\
        \textbf{TF} & \multicolumn{2}{c}{\textbf{ACC}} & \multicolumn{2}{c}{\textbf{ADV-ACC}} & \multicolumn{2}{c}{\textbf{ASR}} & \multicolumn{2}{c}{\textbf{ACC-Drop}} & \multicolumn{2}{c}{\textbf{ACC}} & \multicolumn{2}{c}{\textbf{ADV-ACC}} & \multicolumn{2}{c}{\textbf{ASR}} & \multicolumn{2}{c}{\textbf{ACC-Drop}} \\
        \textbf{proxy} & \textbf{SPR} & \textbf{KT} & \textbf{SPR} & \textbf{KT} & \textbf{SPR} & \textbf{KT} & \textbf{SPR} & \textbf{KT} & \textbf{SPR} & \textbf{KT} & \textbf{SPR} & \textbf{KT} & \textbf{SPR} & \textbf{KT} & \textbf{SPR} & \textbf{KT} \\ \hline
        Head importance & -0.503 & -0.340 & -0.257 & -0.170 & -0.286 & -0.426 & -0.256 & -0.171 & -0.542 & -0.396 & -0.361 & -0.242 & -0.461 & -0.312 & -0.540 & -0.368 \\
        Synaptic Diversity & -0.323 & -0.204 & -0.157 & -0.102 & -0.273 & -0.172 & -0.190 & -0.118 & -0.454 & -0.306 & -0.323 & -0.216 & -0.371 & -0.245 & -0.443 & -0.294 \\
        Synaptic Saliency & 0.361 & 0.248 & 0.175 & 0.119 & 0.317 & 0.221 & 0.156 & 0.104 & -0.133 & -0.080 & -0.117 & -0.079 & -0.078 & -0.045 & -0.123 & -0.072 \\
        Attention confidence & 0.567 & 0.393 & 0.112 & 0.075 & 0.621 & 0.435 & 0.487 & 0.335 & 0.577 & 0.404 & 0.405 & 0.274 & 0.502 & 0.349 & 0.559 & 0.392 \\
        Jacobian Cosine & 0.688 & 0.500 & 0.457 & 0.317 & 0.535 & 0.377 & 0.234 & 0.161 & 0.353 & 0.245 & 0.308 & 0.209 & 0.271 & 0.185 & 0.318 & 0.218 \\
        Head Softmax & 0.760 & 0.563 & 0.243 & 0.160 & 0.756 & 0.567 & 0.543 & 0.383 & 0.681 & 0.494 & 0.466 & 0.317 & 0.595 & 0.419 & 0.668 & 0.479 \\
        \#params & 0.855 & 0.662 & 0.435 & 0.299 & 0.724 & 0.537 & 0.395 & 0.280 & 0.624 & 0.444 & 0.332 & 0.222 & 0.580 & 0.411 & 0.656 & 0.471 \\
        Activation Distance & 0.883 & 0.702 & 0.447 & 0.307 & 0.751 & 0.562 & 0.414 & 0.294 & 0.621 & 0.441 & 0.359 & 0.241 & 0.562 & 0.396 & 0.638 & 0.454 \\
        Ours & 0.903 & 0.728 & 0.446 & 0.304 & 0.778 & 0.588 & 0.440 & 0.312 & 0.638 & 0.456 & 0.364 & 0.245 & 0.582 & 0.413 & 0.660 & 0.473 \\ \hline
        \end{tabular}
        }
    \end{table*}

    \begin{table*}[ht!]
        \centering
        \caption{The four metrics tested on the SST-2 and QNLI adversarial samples obtained using the TB method are used as the SPR and KT between the performance and 9 different proxies.}
        \resizebox{\textwidth}{!}{
        \begin{tabular}{ccccccccccccccccc}
        \hline
        \textbf{Datasets} & \multicolumn{8}{c}{\textbf{SST-2}} & \multicolumn{8}{c}{\textbf{QNLI}} \\
        \textbf{TB} & \multicolumn{2}{c}{\textbf{ACC}} & \multicolumn{2}{c}{\textbf{ADV-ACC}} & \multicolumn{2}{c}{\textbf{ASR}} & \multicolumn{2}{c}{\textbf{ACC-Drop}} & \multicolumn{2}{c}{\textbf{ACC}} & \multicolumn{2}{c}{\textbf{ADV-ACC}} & \multicolumn{2}{c}{\textbf{ASR}} & \multicolumn{2}{c}{\textbf{ACC-Drop}} \\
        \textbf{proxy} & \textbf{SPR} & \textbf{KT} & \textbf{SPR} & \textbf{KT} & \textbf{SPR} & \textbf{KT} & \textbf{SPR} & \textbf{KT} & \textbf{SPR} & \textbf{KT} & \textbf{SPR} & \textbf{KT} & \textbf{SPR} & \textbf{KT} & \textbf{SPR} & \textbf{KT} \\ \hline
        Head importance & -0.507 & -0.343 & 0.157 & 0.093 & -0.421 & -0.277 & -0.373 & -0.242 & -0.513 & -0.354 & 0.352 & 0.235 & -0.552 & -0.382 & -0.599 & -0.418 \\
        Synaptic Diversity & -0.328 & -0.207 & 0.058 & 0.022 & -0.241 & -0.145 & -0.202 & -0.118 & -0.425 & -0.286 & 0.341 & 0.226 & -0.463 & -0.310 & -0.504 & -0.340 \\
        Synaptic Saliency & 0.357 & 0.245 & -0.325 & -0.221 & 0.417 & 0.295 & 0.422 & 0.297 & -0.114 & -0.068 & 0.201 & 0.132 & -0.136 & -0.080 & -0.153 & -0.089 \\
        Attention confidence & 0.568 & 0.395 & -0.554 & -0.384 & 0.625 & 0.438 & 0.638 & 0.448 & 0.549 & 0.383 & -0.341 & -0.224 & 0.595 & 0.419 & 0.634 & 0.450 \\
        Jacobian Cosine & 0.689 & 0.502 & -0.266 & -0.186 & 0.625 & 0.447 & 0.569 & 0.400 & 0.335 & 0.232 & -0.267 & -0.175 & 0.365 & 0.255 & 0.390 & 0.274 \\
        Head Softmax & 0.760 & 0.563 & -0.631 & -0.456 & 0.799 & 0.604 & 0.796 & 0.601 & 0.651 & 0.467 & -0.375 & -0.248 & 0.693 & 0.507 & 0.741 & 0.550 \\
        \#params & 0.855 & 0.666 & -0.516 & -0.378 & 0.857 & 0.665 & 0.818 & 0.620 & 0.606 & 0.431 & -0.361 & -0.246 & 0.652 & 0.470 & 0.706 & 0.516 \\
        Activation Distance & 0.883 & 0.702 & -0.522 & -0.380 & 0.877 & 0.688 & 0.836 & 0.638 & 0.596 & 0.422 & -0.377 & -0.257 & 0.646 & 0.467 & 0.700 & 0.512 \\
        Ours & 0.902 & 0.729 & -0.528 & -0.383 & 0.892 & 0.707 & 0.850 & 0.665 & 0.615 & 0.439 & -0.374 & -0.254 & 0.663 & 0.482 & 0.718 & 0.528 \\ \hline
        \end{tabular}
        }
        \label{table:TB-SST2}
    \end{table*}

    \begin{table*}[ht!]
        \centering
        \caption{The four metrics tested on the SST-2 and QNLI adversarial samples obtained using the PWWs method are used as the SPR and KT between the performance and 9 different proxies.}
        \resizebox{\textwidth}{!}{
        \begin{tabular}{ccccccccccccccccc}
        \hline
        \textbf{Datasets} & \multicolumn{8}{c}{\textbf{SST-2}} & \multicolumn{8}{c}{\textbf{QNLI}} \\
        \textbf{PWWs} & \multicolumn{2}{c}{\textbf{ACC}} & \multicolumn{2}{c}{\textbf{ADV-ACC}} & \multicolumn{2}{c}{\textbf{ASR}} & \multicolumn{2}{c}{\textbf{ACC-Drop}} & \multicolumn{2}{c}{\textbf{ACC}} & \multicolumn{2}{c}{\textbf{ADV-ACC}} & \multicolumn{2}{c}{\textbf{ASR}} & \multicolumn{2}{c}{\textbf{ACC-Drop}} \\
        \textbf{proxy} & \textbf{SPR} & \textbf{KT} & \textbf{SPR} & \textbf{KT} & \textbf{SPR} & \textbf{KT} & \textbf{SPR} & \textbf{KT} & \textbf{SPR} & \textbf{KT} & \textbf{SPR} & \textbf{KT} & \textbf{SPR} & \textbf{KT} & \textbf{SPR} & \textbf{KT} \\ \hline
        Head importance & -0.503 & -0.341 & 0.213 & 0.138 & -0.414 & -0.274 & -0.372 & -0.242 & -0.494 & -0.339 & -0.060 & -0.040 & -0.479 & -0.323 & -0.604 & -0.415 \\
        Synaptic Diversity & -0.324 & -0.205 & 0.065 & 0.031 & -0.234 & -0.139 & -0.196 & -0.112 & -0.406 & -0.271 & -0.082 & -0.055 & -0.373 & -0.244 & -0.489 & -0.323 \\
        Synaptic Saliency & 0.359 & 0.247 & -0.462 & -0.324 & 0.440 & 0.310 & 0.457 & 0.322 & -0.101 & -0.060 & -0.052 & -0.037 & -0.053 & -0.027 & -0.124 & -0.068 \\
        Attention confidence & 0.563 & 0.391 & -0.704 & -0.507 & 0.657 & 0.465 & 0.688 & 0.489 & 0.529 & 0.368 & 0.073 & 0.047 & 0.530 & 0.370 & 0.626 & 0.444 \\
        Jacobian Cosine & 0.689 & 0.501 & -0.354 & -0.244 & 0.432 & 0.432 & 0.551 & 0.385 & 0.319 & 0.220 & 0.114 & 0.080 & 0.279 & 0.191 & 0.359 & 0.248 \\
        Head Softmax & 0.757 & 0.561 & -0.796 & -0.602 & 0.828 & 0.635 & 0.840 & 0.648 & 0.629 & 0.448 & 0.090 & 0.061 & 0.628 & 0.447 & 0.745 & 0.549 \\
        \#params & 0.854 & 0.665 & -0.742 & -0.555 & 0.888 & 0.705 & 0.873 & 0.683 & 0.590 & 0.419 & -0.006 & -0.005 & 0.632 & 0.454 & 0.749 & 0.553 \\
        Activation Distance & 0.882 & 0.701 & -0.742 & -0.555 & 0.905 & 0.729 & 0.885 & 0.699 & 0.577 & 0.407 & -0.005 & -0.005 & 0.609 & 0.434 & 0.732 & 0.535 \\
        Ours & 0.901 & 0.727 & -0.739 & -0.550 & 0.914 & 0.742 & 0.890 & 0.706 & 0.597 & 0.425 & 0.006 & 0.004 & 0.631 & 0.453 & 0.753 & 0.557 \\ \hline
        \end{tabular}
        }
    \label{table:PWWs-SST2}
    \end{table*}

    \begin{figure*}[ht]
        \begin{minipage}{0.24\linewidth}
            \centering
            \includegraphics[width=0.9\textwidth]{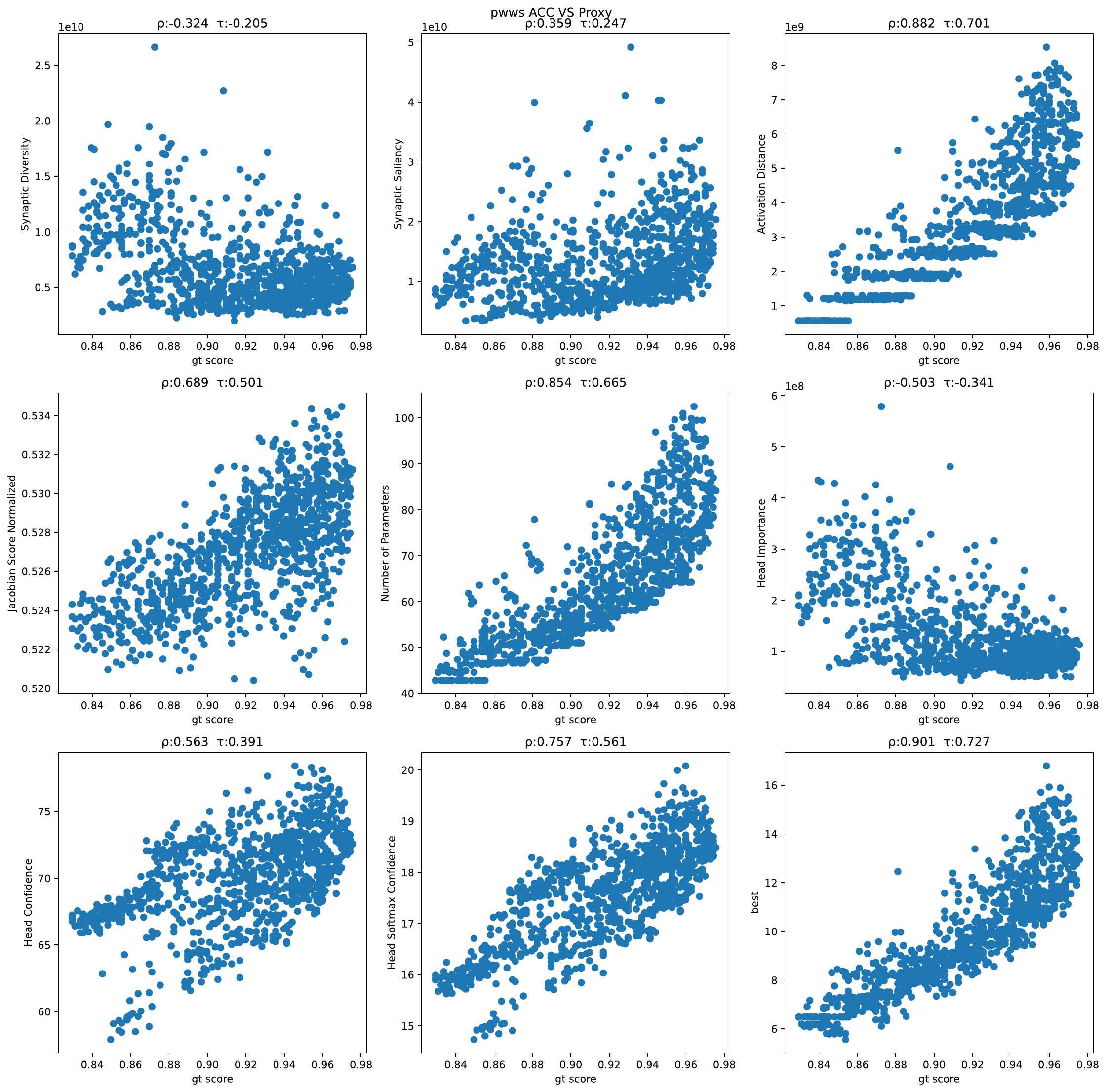}
            \caption*{(a)}
        \end{minipage}
        \begin{minipage}{0.24\linewidth}
            \centering
            \includegraphics[width=0.9\textwidth]{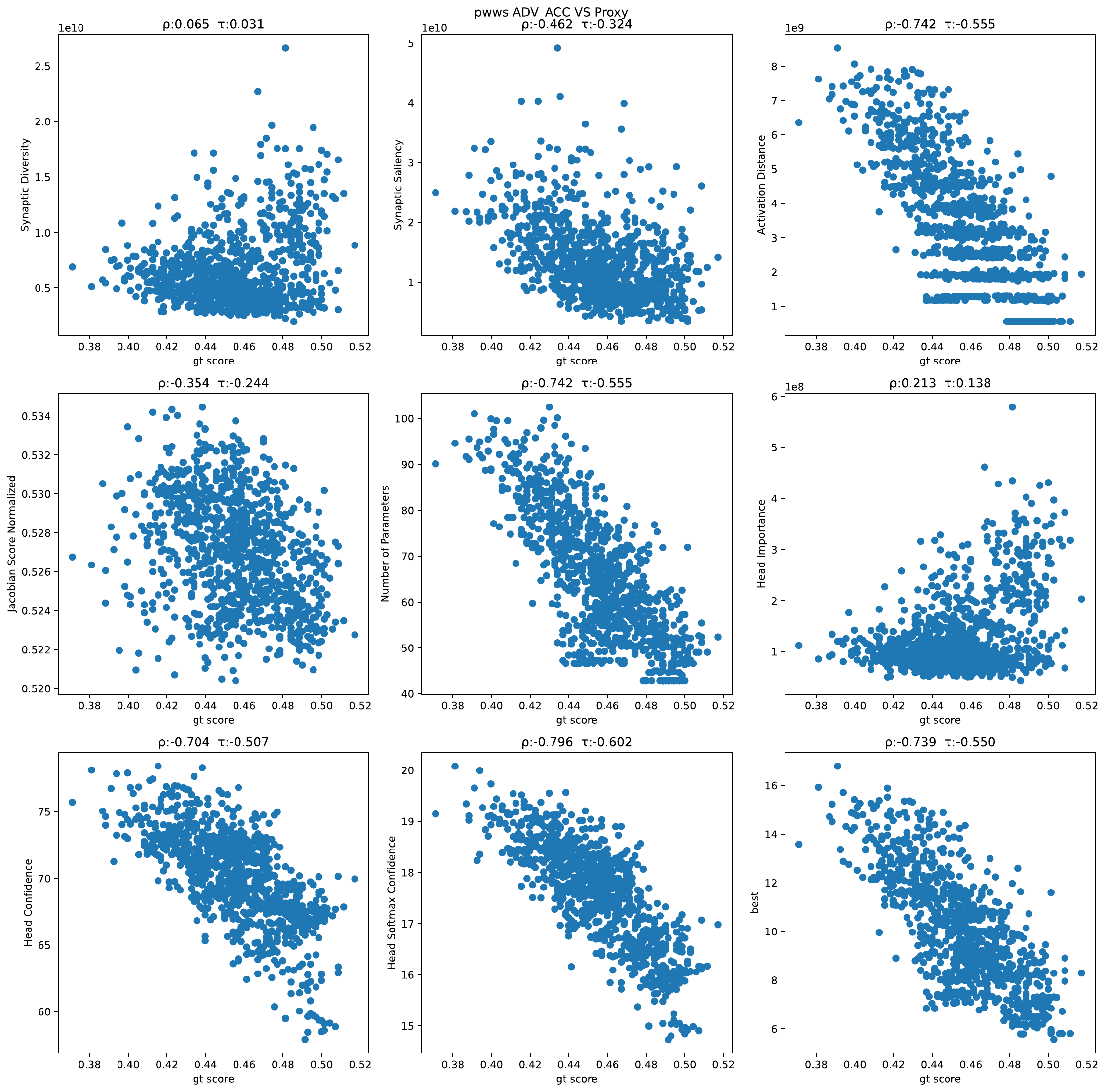}
            \caption*{(b)}
        \end{minipage}
        \begin{minipage}{0.24\linewidth}
            \centering
            \includegraphics[width=0.9\textwidth]{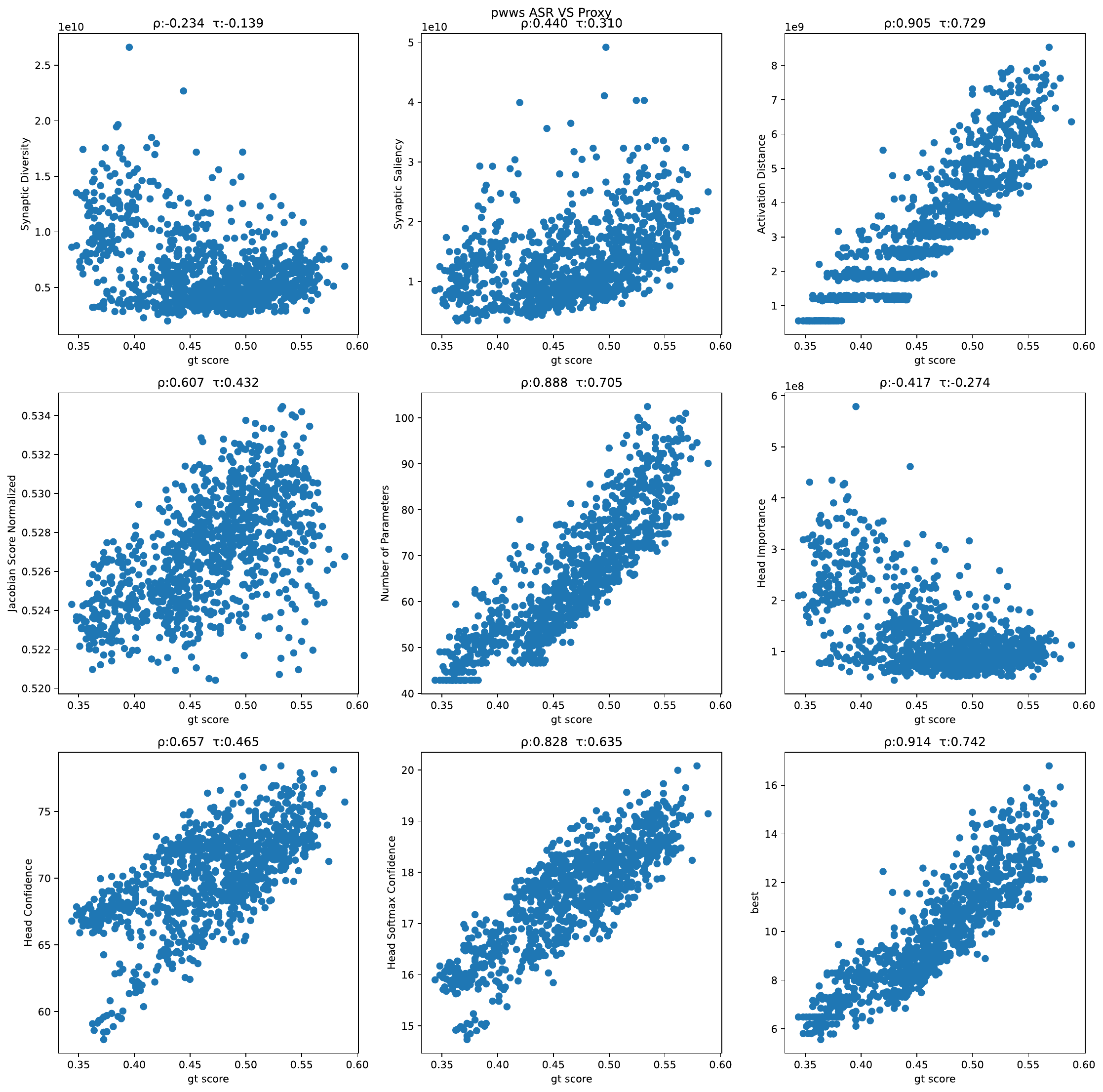}
            \caption*{(c)}
        \end{minipage}
        \begin{minipage}{0.24\linewidth}
            \centering
            \includegraphics[width=0.9\textwidth]{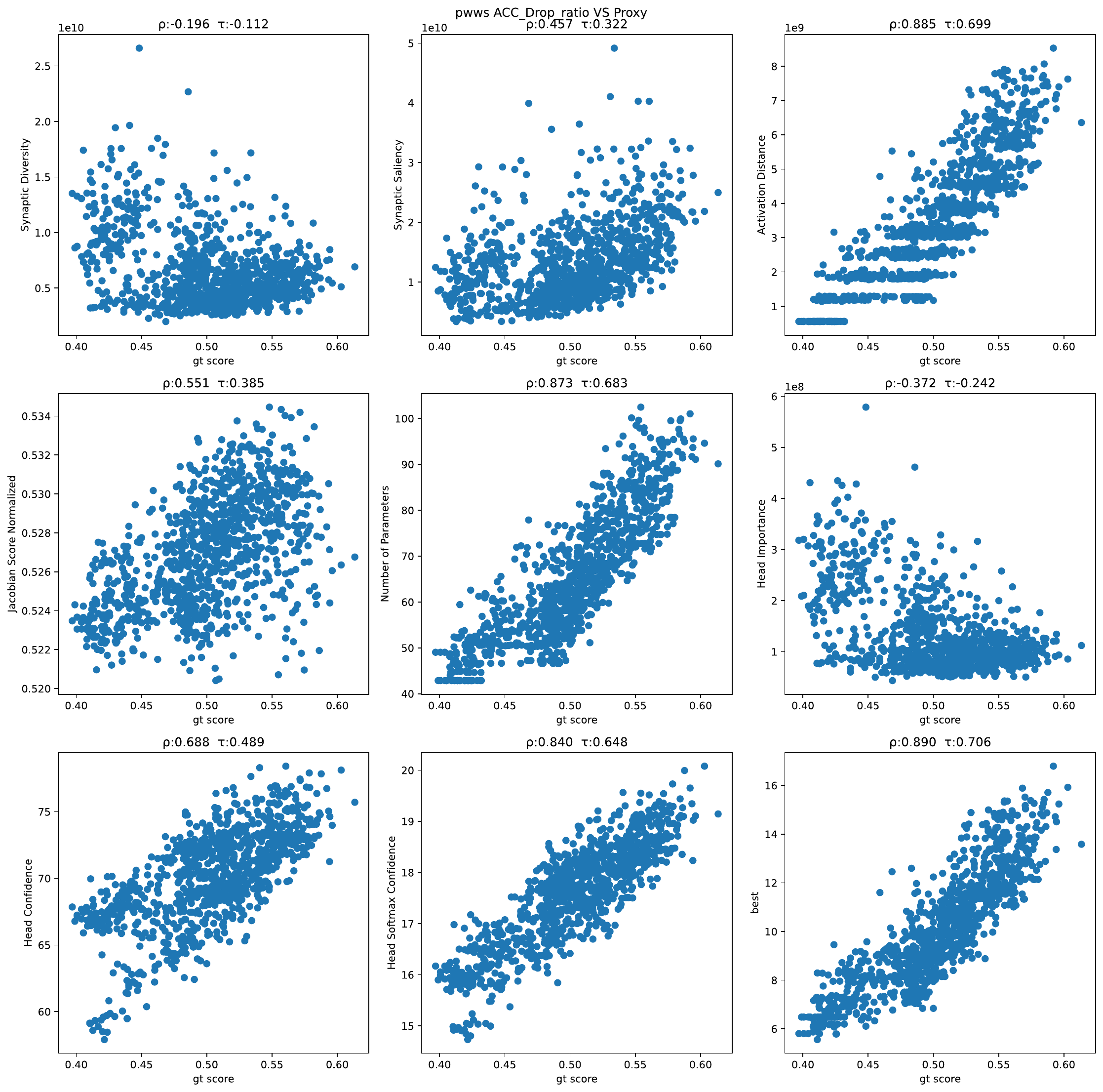}
            \caption*{(d)}
        \end{minipage}
        \caption{Figure shows a scatter plot of the ranking relationship between the proxy and four different metrics in PWWS-SST-2, figure (a) is proxies vs. ACC, figure (b) is proxies vs. ADV-ACC, figure (c) is proxies vs. ASR, figure (d) is proxies vs. ACC-Drop.}
        \label{fig:Pwws Scatter plot SST2}
    \end{figure*}

    \begin{figure*}[ht]
        \centering
        \includegraphics[width=1\textwidth]{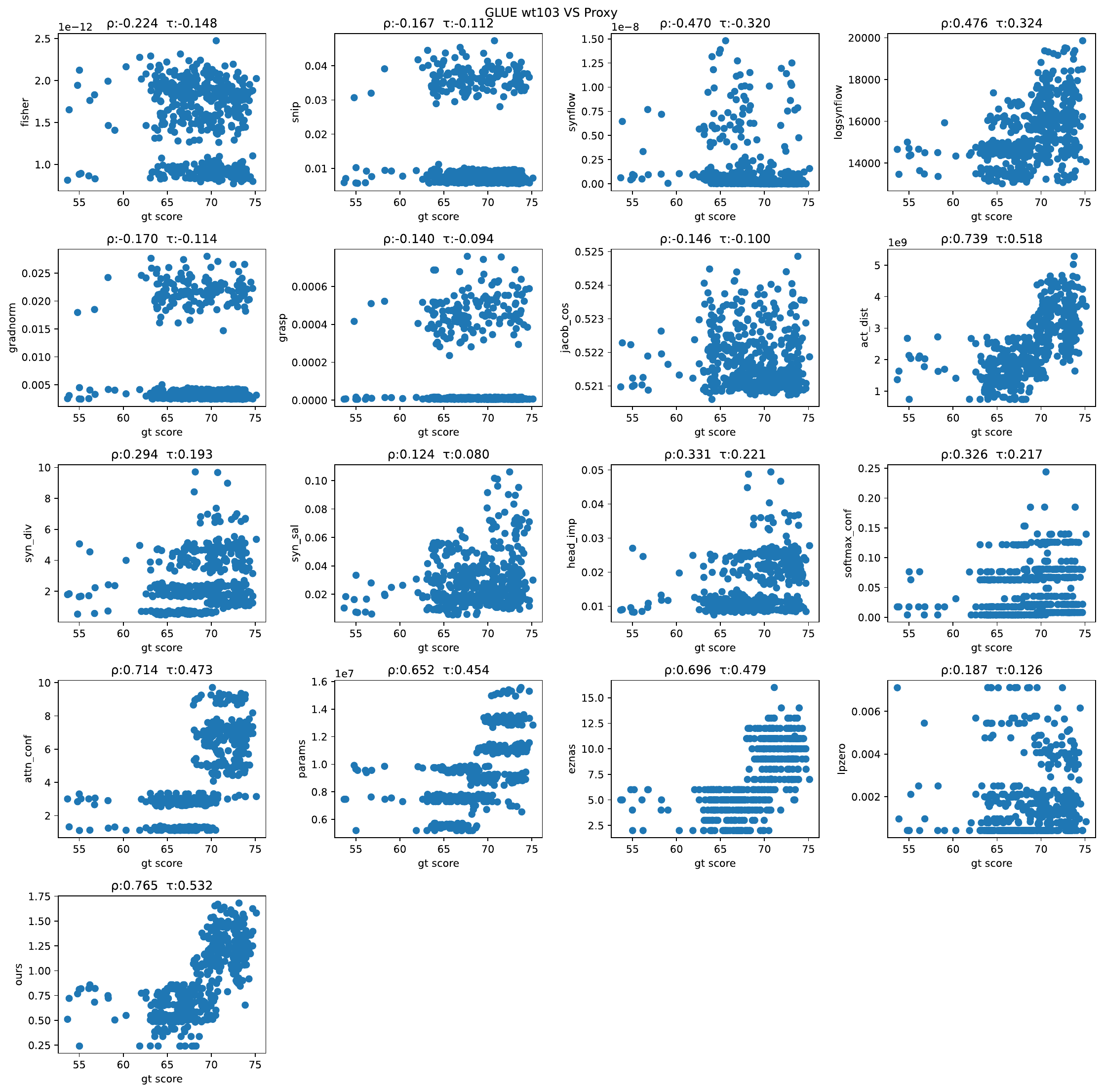}
        \caption{Correlation of training-free proxies, pruning metrics as proxies ranking with GLUE Ranking on 500 architectures randomly sampled in FlexiBERT benchmark. }
        \label{fig:plot WT103 flexibert}
    \end{figure*}

\end{document}